\documentclass[11pt]{article}
\usepackage{latexsym}
\usepackage{color}
\usepackage{graphicx, amsmath}
\usepackage{wrapfig}
\usepackage{subfigure}
\usepackage{graphicx}
\usepackage{url}
\usepackage{CJK,graphicx}
\usepackage{algorithmic}
\usepackage{multirow}
\usepackage{subfigure}
\usepackage{float}
\usepackage{appendix}
 \usepackage{textcomp, libertine}

\usepackage{lineno}
\usepackage{flushend}
\usepackage{pifont}

\usepackage{amsthm}
\usepackage{graphics,epsfig,graphicx,subfigure}
\usepackage[table]{xcolor}
\usepackage{helvet,epsfig,graphicx,subfigure}
\usepackage[table]{xcolor}

\usepackage{courier}
\usepackage{lineno}
\usepackage{flushend}
\usepackage{epsfig,graphicx,subfigure}
\usepackage{booktabs}
\usepackage{CJK,graphicx}
\usepackage{algorithm}
\usepackage{color}
\usepackage{fancyhdr}
\usepackage{hyperref}

\usepackage{textcomp}

\usepackage{tcolorbox}

\usepackage[normalem]{ulem}

\theoremstyle{definition}

\theoremstyle{remark}

\numberwithin{equation}{section}

\usepackage{algorithm}
\usepackage{tikz}
\usepackage{amsmath} \renewcommand{\algorithmicrequire}{\textbf{Input:}} \renewcommand{\algorithmicensure}{\textbf{Output:}}
\usepackage{booktabs,array,xcolor,colortbl,multirow,epsfig,graphicx,subfigure,amsmath,amsthm,amsfonts,amssymb,bm}

\begin{document}

\title{\Large HMRL: Hyper-Meta Learning for Sparse Reward Reinforcement Learning Problem}

\author{Yun~Hua\thanks{School of Computer Science and Technology, East China Normal University, Shanghai 200092, China. (E-mail: 52194501002@stu.ecnu.edu.com)}
\and Xiangfeng~Wang\thanks{School of Computer Science and Technology, East China Normal University, Shanghai 200062, China. (E-mail: xfwang@sei.ecnu.edu.cn)}
\and Bo~Jin\thanks{School of Computer Science and Technology, East China Normal University, Shanghai 200062, China. (E-mail: bjin@cs.ecnu.edu.cn)}
\and Wenhao~Li\thanks{School of Computer Science and Technology, East China Normal University, Shanghai 200092, China. (E-mail: 52194501026@stu.ecnu.edu.cn)}
\and Junchi~Yan\thanks{Department of Computer Science and Engineering, Artificial Intelligence Institute, Shanghai Jiao Tong University, Shanghai 200240, China. (E-mail: yanjunchi@sjtu.edu.cn)}
\and Xiaofeng~He\thanks{School of Computer Science and Technology, East China Normal University, Shanghai 200062, China. (E-mail:xfhe@sei.ecnu.edu.cn)}
\and Hongyuan~Zha\thanks{School of Computational Science and Engineering, College of Computing, Georgia Institute of Technology, USA. (E-mail: zha@cc.gatech.edu)}
}
\date{}

\maketitle

\begin{abstract}
In spite of the success of existing meta reinforcement learning methods, they still have difficulty in learning a meta policy effectively for RL problems with sparse reward. In this respect, we develop a novel meta  reinforcement learning framework called Hyper-Meta RL(HMRL), for sparse reward RL problems. It is consisted with three modules including the cross-environment meta state embedding module which constructs a common meta state space to adapt to different environments; the meta state based environment-specific meta reward shaping which effectively extends the original sparse reward trajectory by cross-environmental knowledge complementarity and as a consequence the meta policy achieves better generalization and efficiency with the shaped meta reward. Experiments with sparse-reward environments show the superiority of HMRL on both transferability and policy learning efficiency.
\end{abstract}

\section{Introduction}
Deep reinforcement learning (RL) have shown great success in games \cite{silver2016mastering, mnih2015human-level,tessler2016a}, and also practical areas like robotics \cite{duan2016benchmarking} and traffic light management \cite{wei2018intellilight}. To achieve generalization  and efficiency on multiple tasks, meta reinforcement learning (meta-RL), which combines RL and meta learning, has recently been studied by extracting meta knowledge to help efficient policy learning on new tasks.
The works \cite{wang2016learning,duan2016rl} model task embedding as meta knowledge, to extract task features. Direct attention is applied to policy learning in~\cite{finn2017model}, which is extended in~\cite{xu2018meta,gupta2018meta,lan2019meta,gurumurthy2019mame} by using basic components in RL as meta knowledge, e.g. hyper-parameters, return function, advantage function and etc.

These methods are still algorithmically limited, especially with sparse reward. Some meta-RL methods focus on the reward sequence: PEMIRL \cite{yu2019meta} introduces meta inverse RL to directly learn the reward signals. NoRML \cite{yang2019norml} learns meta advantage values as additional rewards. \cite{zou2019reward} employs meta-RL to design reward shaping function, to speed up training on new tasks. The sparse reward problem has been partially addressed by such methods, while it is still far from resolved.
Existing meta-RL models mostly deal with tasks within one environment. The setting for transfer across different environments and the relevant techniques have not been well studied in literature, especially with sparse rewards. In particular, we make two observations: 1) The rewards of tasks in the same environment tend to be homogeneous which may weaken the capacity for effective cross-task reward utilization; 2) While cross-environment meta knowledge can be complementary to enrich the reward over the tasks across environments, thus mitigates the sparse reward issue.

Motivated by these observations, this work aims to devise a cross-environment meta learning approach with sparse reward. We propose a hyper meta RL framework, called {\em{Hyper-Meta RL}}, which consists of three meta modules. We devise the cross-environment meta state embedding module and environment-specific meta reward shaping module to support the general meta policy learning module. The meta state is embedded across environments. Based on the general meta-state, the meta reward shaping module is dedicated to generate targeted reward shaping function. Further, the meta state and additional reward signals can improve the meta policy model to reduce the negative influence of missing reward signal. 
Moreover, the shared meta state space benefits meta reward shaping and provides a flexible mechanism to integrate domain knowledge from multiple environments.
The Hyper-Meta RL framework is illustrated in Figure~\ref{fig:framework}. The main contributions are:
1)~For the sparse reward setting, a hyper-meta RL approach is proposed, to jointly utilize the information of tasks from multiple environments. This is in contrast to existing works for a single environment.
As a natural way to improve generalization for meta policy learning, 
this proposes a novel meta-RL method on utilizing cross-environment information for sparse reward RL problems;

2)~For each environment, an environment-specific meta reward shaping technique is developed. The shaping is fed with the meta state learned from domain knowledge in both the current and other environments. It can speed up new task learning either with or without meta policy, and meanwhile the policy consistency can be theoretically ensured;

3)~We show that our methods can effectively improve both generalization capability and transfer efficiency, especially under environments with notable heterogeneity, as shown in our experiments.

Finally, we point out that our approach in fact can be interpreted as a hierarchical structure for task learning. The environment acts as the group of tasks, which can either be formed by prior knowledge or other means. In contrast, existing methods treat all tasks as in a single environment. Hence the improvement can be attributed to both our hyper-meta methods as well as the prior of the task set structure. At least, our method allows a direct and effective way to incorporate environment priors, which are often readily available or naturally arise.
Furthermore, our method don't need too much computing resources because the only additive parts which need external computing resources are environment-specific meta reward shaping and cross-environment meta state embedding and they only need some simple neural networks.
\section{Related work}
\paragraph{Sparse Reward Reinforcement Learning}
As the reward signal is the only supervised information in reinforcement learning, sparse reward problems will cause many negative influence on the training of reinforcement learning. Sparse reward reinforcement learning methods contain these classes: 1) Reward shaping or reward design~\cite{ng1999policy,riedmiller2018learning,hare2019dealing} methods; 2) Curriculum learning methods~\cite{narvekar2016source,svetlik2017automatic,florensa2017reverse,wohlke2020performance}; 3) Bayesian reinforcement learning~\cite{ross2007bayes,ghavamzadeh2016bayesian}. The reward shaping or reward design methods directly learns additional reward to make the sparse reward dense while curriculum learning transfer policy from easier task as a prior and Bayesian reinforcement learning learns prior information about tasks.

\paragraph{Meta Reinforcement Learning.}
It aims to learn a common structure between different tasks to solve new and relevant tasks efficiently. Meta RL methods can also be divided into two categories: model-based meta RL and model-free meta RL. The model-based methods have the advantage to directly model dynamic environments, specifically by the environment parameters \cite{clavera2018learning,saemundsson2018meta,arcari2020meta} or priors \cite{fu2016one}. 

Model-free meta RL methods can be divided into these classes: 1) recurrence based meta-RL methods, which use recurrent neural network (RNN) to establish the meta task embedding \cite{duan2016rl,mishra2017simple,wang2016learning}; 2) gradient based meta-RL methods, which are mostly extended from MAML \cite{finn2017model} aim to learn a meta policy or some well-defined meta parameters simultaneously. The meta information can be set as the exploration strategy \cite{gurumurthy2019mame,gupta2018meta,zhang2020learn}, advantage function \cite{yang2019norml}, reward shaping \cite{zou2019reward}, return function~\cite{xu2018meta} and etc; 3) hybrid meta-RL methods~\cite{lan2019meta,sung2017learning},  considered as the hybrid of the above two techniques.
\cite{gurumurthy2019mame,gupta2018meta} set the exploration policy as the meta knowledge, while \cite{yang2019norml} meta-learns the advantage function instead.

These meta-RL methods mostly focus on RL with dense reward, and only a few pay attention to sparse reward case.
NoRML \cite{yang2019norml} meta-learns the advantage function for sparse reward tasks; \cite{zou2019reward} introduces MAML to enhance the reward shaping technique, to efficiently handle sparse reward
;VariBAD~\cite{zintgraf2020varibad} uses the Bayesian formulation to learn prior information to solve sparse reward problem in meta RL. Above meta-RL methods mostly address learning of different tasks from the same environment; DREAM~\cite{liu2020explore} build a new exploration stricture to solve no reward problem only in adapting but not training time. For a more general meta-RL problem with sparse reward, utilizing limited reward signals from relevant environments to design the meta-RL algorithm can be a \textbf{natural but currently under-investigated way} to increase the generalization and transfer abilities.

\section{Hyper-Meta RL with Sparse Reward}
Our key idea is to combine model-based meta RL and model-free meta RL. In our setting, tasks are sampled from different distributions based on environments, and environments are sampled from a fixed distribution. Instead of modeling the environment directly, our approach indirectly utilizes the common knowledge to model the cross-environment meta knowledge, which is denoted as the {\em{Cross-Environment Meta State Embedding}}.
\begin{figure*}[tb!]
    \begin{center}
    \centerline{\includegraphics[width=1\textwidth]{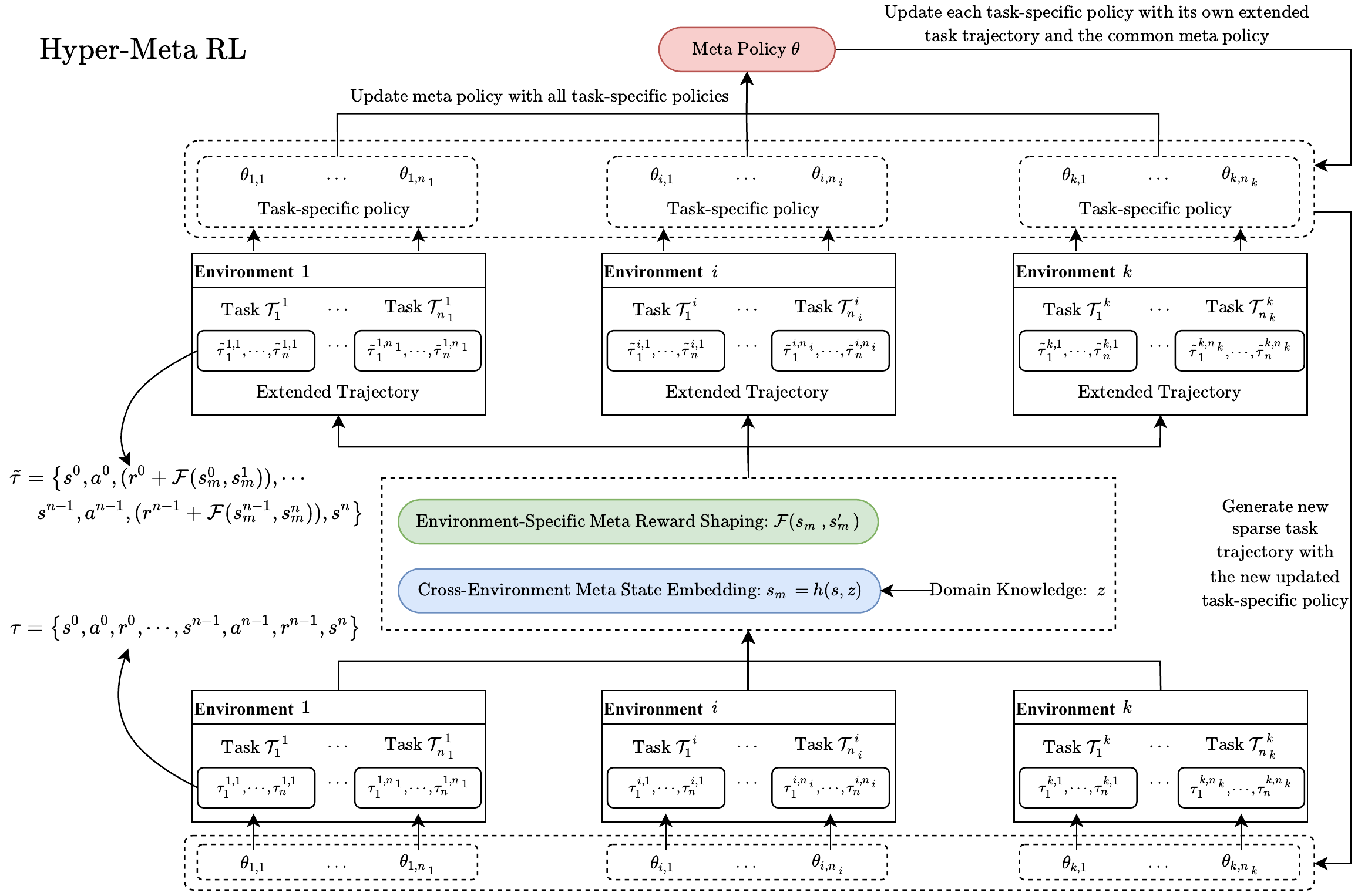}}
    \caption{Overview of HMRL: 1) Tasks from $k$ environments generate trajectories by their task specific policies $\{ \theta_{i,j} \}$ separately; 2) {\em{Cross-Environment Meta State Embedding}} (Meta-SE) module obtains meta states $s_m$ from domain knowledge in multi-environments and generated trajectories. 3) {\em{Environment-specific Meta Reward Shaping}} (Meta-RS) module employs the meta states obtained by Meta-SE to calculate the meta reward shaping, as to generate extended trajectories. 4) Meta Policies (Meta-PC) $\theta$ are learned based on the extended trajectories with meta state embedding and meta reward shaping. Environments can be formed by prior or by other means in advance. The dashed box part is the cross-environment part in HMRL.}
    \label{fig:framework}
    \end{center}
\end{figure*}
As detailed in Figure~\ref{fig:framework}, our Hyper-Meta RL method consists of three modules:  1) cross-environment meta state embedding module; 2) environment-specific meta reward shaping module; 3) meta policy learning module. In the following part of this section, we will first introduce the preliminaries and problem formulation, then we will introduce the three modules. Further we will prove the policy consistency to maintain on new tasks thus the method can be extended to different tasks in a stable fashion.

\subsection{Preliminaries and Problem Formulation}
We first introduce the popular meta-learning method Model-Agnos-\\tic Meta-Learning (MAML) in details, which serves as a basic technique employed in our approach. MAML meta-learns an policy initialization $\theta$ (meta policy) of model parameters based on a series of tasks $\left\{{\mathcal{T}}_i\right\}$ by:
\begin{equation}\label{eq:maml-iteration}
    \left\{\begin{array}{ll}
        \theta_i \!\!&\longleftarrow \theta - \alpha \nabla_\theta {\mathcal{L}}_{{\mathcal{T}}_i}(\theta),\\
        \theta \!\!&\longleftarrow \theta - \beta \nabla_\theta \sum_{{\mathcal{T}}_i} {\mathcal{L}}_{{\mathcal{T}}_i}(\theta_i),
        \end{array}\right.
\end{equation}where $\theta_i$ is the task-specific policy with respect to task ${\mathcal{T}}_i$, and is updated from the meta policy $\theta$; 
$\alpha$ and $\beta$ are both learning rates;
${\mathcal{L}}_{{\mathcal{T}}_i}$ denotes the loss with respect to task ${\mathcal{T}}_i$.

Now we consider a set of tasks $\left\{ {\mathcal{E}}_1,{\mathcal{E}}_2,\cdots,{\mathcal{E}}_k \right\}$ from $k$ environments, where ${\mathcal{E}}_i$ is the task subset for environment $i$. The extended Markov decision process (MDP) is defined:
$$
{\mathcal{E}}_i = \left( \left\{ {\mathcal{S}}_j^i\right\}_{j=1}^{n_i}, {\mathcal{S}}_{m}, \left\{ {\mathcal{A}}_j^i\right\}_{j=1}^{n_i}, \left\{ {\mathcal{P}}_j^i \right\}_{j=1}^{n_i}, {\mathcal{P}}_{m}, \gamma, \hat{\mathcal{R}} \right),
$$
where there are $n_i$ tasks for environment $i$; ${\mathcal{S}}_j^i$, ${\mathcal{A}}_j^i$ denote the state and action set of the task $j$ for environment $i$ respectively, and ${\mathcal{P}}_j^i(s'|s,a)$ denotes the corresponding transition probability function with $s,s'\in {\mathcal{S}}_j^i$, $a\in {\mathcal{A}}_j^i$;
a shared meta state space ${\mathcal{S}}_{m}$ is introduced based on some domain knowledge of all environments, and the corresponding meta transition probability function is denoted as ${\mathcal{P}}_{m} (s'|s,a)$; $\gamma$ denotes the discount factor; our newly devised reward $\hat{\mathcal{R}}$ is used as will be discussed later in the following.

\medskip
\noindent{\em{Reward Shaping}}: Further we introduce the reward shaping. The objective in RL is to maximize expected discounted long-term return:
$$
\mathbb{E} \left[ G:= \sum_{t=0}^{T} \gamma^t r_t \right].
$$The reward signal is often weak or even missing in many applications~\cite{ng1999policy,riedmiller2018learning,hare2019dealing}, making it more difficult to learn the optimal policy. The reward shaping technique is used to solve this problem by adding an extra reward signal ${\mathcal{F}}$, leading to the composite reward ${\mathcal{R}}_{\mathcal{F}} = {\mathcal{R}}+{\mathcal{F}}$. Usually, the shaping reward ${\mathcal{F}}$ encodes some heuristic knowledge for more effective optimal policy search. It is worth noting that modifying the original reward will eventually change the task and may cause the policy inconsistency. To solve this problem, potential-based reward shaping is proposed in \cite{ng1999policy}:
\begin{equation}
    {\mathcal{F}}(s,a,s')=\gamma \phi(s')-\phi(s),
\end{equation}
where $s$, $s'$ denote two consecutive states and $a$ denotes the action; $\phi$ is a potential function; the function value ${\mathcal{F}}$ denotes the reward shaping signal as the difference between the potential of $s'$ and $s$; $\gamma$ denotes a discount rate.
The dynamic potential-based reward shaping in \cite{devlin2012dynamic} extends the original reward shaping to allow dynamic shaping and maintain the policy consistency:
\begin{equation}
{\mathcal{F}}(s, t, s', t')=\gamma \phi(s',t') - \phi(s,t),
\end{equation}
where $t$ and $t'$ are two consecutive time stamps.

\subsection{Cross-Environment Meta State Embedding}
\label{subsec:cross_env}
The cross-environment meta state embedding module focuses on embedding both the domain knowledge of the environment and the trajectory of the environment-specific tasks into a unified meta state space.
The prior knowledge among different environments should be properly determined to define the shared meta state space ${\mathcal{S}}_m$. 
For example, the meta state can be defined as the agent coordination in both 1st person game and 3rd person game because the agent coordination is in a same space so that it can be shared.
The setting of the meta state space is flexible, while the embedding function can further be pertinently learned in the closed loop of the whole algorithm.
We can easily use some shared information in different environments' state spaces, and extract them into the shared meta state space.
For the Hallway problem, the concatenation of the location information is established as the meta state embedding function $h$ without learning.
More generally, the function $h$ can directly embed the original agent state together with some global features of the related task into the common meta state space (or a shared latent space).
For the Maze problem, we employ a linear function parameterized by each environment's scenario size (domain knowledge $z$) as the state embedding function $h$, which makes a common mini-map as the common meta state.

\subsection{Environment-specific Meta Reward Shaping}
\label{subsec:each_env}
The environment-specific meta reward shaping module focuses on constructing the shaped reward, i.e., $\hat{\mathcal{R}} = \mathcal{R} + \mathcal{F}$.
This module is based on the the potential-based reward shaping, while the cross-environment meta states are set to be the input of the potential function $\phi$.
The formulation is as:
\begin{equation}\label{eq:potential_based_reward_shaping}
    {\mathcal{F}}(s_{m}, s_{m}') = \gamma \phi(s_{m}') - \phi(s_{m}),
\end{equation}
where $s_{m}\in {\mathcal{S}}_{m}$ and $s'_{m}\in {\mathcal{S}}_m$. The potential function $\phi$ plays the key role in the above reward shaping, which is meta-learned. The parameter $\gamma$ denotes the discount factor of the RL problem. The cross-environment meta sate contains not only the state information but also the environment-specific task information.
The potential value $\phi(s_{m})$ will also be environment driven. Eventually, based on the cross-environment meta state embedding $h(s;z)$ introduced in \ref{subsec:cross_env}, the cross-environment meta states based potential function can be further expressed as:
\begin{equation}\label{eq:potential_based_reward_shaping_h}
    {\mathcal{F}}(s_{m}, s_{m}') = \gamma \phi(h(s';z)) - \phi(h(s;z)).
\end{equation}

So that if we regard the cross environment meta state embedding $h(.;z)$ as part of the potential function $\phi$. Then it is likely to claim that the policy consistency for reward shaping will still holds. And we have proved it in the \ref{subsec:theory}. The policy consistency will strength the adaption to new tasks.

\subsection{Meta Policy Learning}
\label{subsec:meta_policy}
As shown in Fig.~\ref{fig:framework}, the meta policy is learned based on the extended trajectories of different tasks, while the meta state embedding and meta reward shaping are employed to support meta policy implicitly.
We introduce the MAML framework to compute the meta policy iteratively. For the $j$-th task ${\mathcal{T}}_j^i$ of environment ${\mathcal{E}}_i$, the agent aims to find the task-specific policy $\theta_{i,j}$ which can maximize the accumulated shaped reward $\hat{\mathcal{R}}$, and the loss with respect to task ${\mathcal{T}}^i_j$ is:
\begin{equation}\label{eq:loss-rs}
\begin{split}
\!\!\!{\mathcal{L}}_{{\mathcal{T}}^i_j} (\theta_{i,j}) \!\!:=\!\! - {\mathbb{E}}_{s_{t}^{i,j}, a_{t}^{i,j}\sim \theta_{i,j}}
    \left[ \sum_{t=0}^{T} {\gamma^t \hat{{\mathcal{R}}} (s_{t}^{i,j}, a_{t}^{i,j})}  \right],\\
{\hat{{\mathcal{R}}} (s_{t}^{i,j}, a_{t}^{i,j})} ={{\mathcal{R}} (s_{t}^{i,j}, a_{t}^{i,j}) + {\mathcal{F}}(s_{m}^{i,j,t},s_{m}^{i,j,t+1}),}
\end{split}
\end{equation}
where $s_{t}^{i,j}\in {\mathcal{S}}_j^i$ and $a_{t}^{i,j}\in {\mathcal{A}}_j^i$; $s_{m}^{i,j,t+1}\in {\mathcal{S}}_m$ and $s_{m}^{i,j,t}\in {\mathcal{S}}_m$ are both meta sate vectors corresponding to $s_{t+1}^{i,j}$ and $s_{t}^{i,j}$ respectively.
The meta state based meta reward shaping is introduced to modify original reward in order to increase the training efficiency.
Actually each meta state $s_{m}^{i,j,t}$ in ${\mathcal{T}}^i_j$ is determined by the original state $s_{t}^{i,j}$ and some fixed task features, so that the shaping function value ${\mathcal{F}}(s_{m}^{i,j,t},s_{m}^{i,j,t+1})$ is fixed with respect to $\theta_{i,j}$.
The gradient of ${\mathcal{L}}_{\mathcal{T}^i_j}$ corresponding to $\theta_{i,j}$ is the same as follows:
\begin{equation}\label{eq:loss-or}
\tilde{\mathcal{L}}_{{\mathcal{T}}^i_j} (\theta_{i,j}) := - {\mathbb{E}}_{s_{t}^{i,j}, a_{t}^{i,j}\sim \theta_{i,j}}
    \left[ \sum_{t=0}^T {\gamma^t \mathcal{R}} (s_{t}^{i,j}, a_{t}^{i,j}) \right],
\end{equation}which also denotes the original loss function without meta reward shaping.
The meta policy $\theta$ and task-specific policies $\left\{\theta_{i,j}\right\}$ can be iteratively learned by Eq.~\ref{eq:loss-rs}, following MAML.

The meta state embedding $h$ and meta reward shaping $\phi$ will be alternatively updated together with the meta policy.
Before discussing details of learning the meta state embedding $h$ and the meta reward shaping potential function $\phi$, we can assume that for each meta state $s^*_{m}\in {\mathcal{S}}_m$, no less than one original state corresponds to it. Without loss of generality, the state set corresponding to $s^{*}_{m}$ is denoted as ${\mathcal{S}} (s^{*}_{m})$. The potential function $\phi$ can be expressed as:
\begin{equation}\label{eq:potential-func}
\phi (s^{*}_{m}) = {\mathbb{E}}_{\left\{ {\mathcal{T}}^i_j \right\}} \left[ {\mathbb{E}}_{s\in {\mathcal{S}} (s^{*}_{m})} \left( V^*_{{\mathcal{T}}^i_j} (s) \right) \right],
\end{equation}where function $V^*(\cdot)$ denotes the classical optimal value function.
However, the optimal value function $V^*$ is extremely hard to guarantee, we employ the return function to approximately estimate the optimal value function, i.e.,
\begin{equation}\label{eq:potential-func-appro}
\phi (s^{*}_{m}) = {\mathbb{E}}_{\left\{ {\mathcal{T}}^i_j \right\}} \left[ {\mathbb{E}}_{s\in {\mathcal{S}} (s^{*}_{m})} \left( {\rm{Return}}_{{\mathcal{T}}^i_j} (s) \right) \right],
\end{equation}where the return function is calculated as follows:
$${\rm{Return}} (s_t) = \left\{\begin{array}{ll}
                                {\hat{\mathcal{R}}}(s_t) + \gamma {\rm{Return}} (s_{t+1}), & t < T;\\
                                {\hat{\mathcal{R}}}(s_t), & t = T.
                                \end{array}\right.
$$
The loss function with respect to the meta state embedding and meta reward shaping can be established based on Eq.~\ref{eq:potential-func-appro}.
For each $s_m\in {\mathcal{S}}_m$, we define the loss function as follows:
\begin{equation}\label{eq:loss-two-s_m}
{\mathcal{L}}_{s_{m}} := \left( {\mathbb{E}}_{\left\{ {\mathcal{T}}^i_j \right\}} \left[ {\mathbb{E}}_{s\in {\mathcal{S}} (s_{m})} \left( {\rm{Return}}_{{\mathcal{T}}^i_j} (s) \right) \right]  - \phi (s_{m}) \right)^2,
\end{equation}
Further by fixing the last $\phi (s_{m})$ with $h$ and $\phi$ from former iteration, and considering all significant meta state $s_{m} \in S_{m}$ (depending on specific test problems), we  obtain the overall loss for meta state embedding and meta reward shaping:
\begin{equation}\label{eq:loss-two}
{\mathcal{L}} (\phi, h) := \sum_{s_{m}} {\mathcal{L}}_{s_{m}}.
\end{equation}
The meta state embedding function $h$ and meta reward shaping potential function $\phi$ are both updated by gradient descent via Eq.~\ref{eq:loss-two}. The overall meta-RL scheme is shown in Alg. \ref{alg:meta-rl}. With the obtained meta policy $\theta$, meta state embedding $h$ and meta reward shaping $\phi$, we can do fine-tune on new tasks and the detailed algorithm is given in Alg. \ref{alg:fine-tune}. To emphasize, new tasks can utilize only the obtained meta state embedding and meta reward shaping modules, without the meta policy. These two meta modules can independently be applied to new tasks with new policy structure, thus their policy structure can be different from that of the meta policy.

\begin{algorithm}[tb!]
\caption{HMRL: Hyper-Meta $\{{state, reward, policy}\}$ for Reinforcement Learning with Sparse Reward}
\label{alg:meta-rl}
\begin{algorithmic}[1]
   \STATE {\bfseries Input:} Task trajectory set: $\tau({\mathcal{M}})$;
   Distribution over tasks: $p({\mathcal{T}})$; Learning rates: $\alpha, \beta$; 
   \STATE Initialization: $\theta$, $\left\{\theta_j^i\right\}$, $h$, $\phi$;
   \WHILE{not done}
   \STATE Sample batch of environments  $\{\tilde{{\mathcal{E}}}\sim p({\mathcal{E}})\}$
    \FOR{each environments}
    \STATE Sample batch of tasks $\{ \tilde{{\mathcal{T}}}\sim p({\mathcal{T}}|\tilde{{\mathcal{E}}}) \}$;
    \FOR{each $\tilde{{\mathcal{T}}}$}
        \STATE Sample $m$ extended trajectories ${\mathcal{D}} := \{ \tau'(\tilde{{\mathcal{T}}}) \}$ with meta states $h$, reward shaping $\phi$, policy $\theta$;
        \STATE Evaluate $\nabla_{\theta} {\mathcal{L}}_{\tilde{\mathcal{T}}}(\theta)$ using ${\mathcal{D}}$ and Eq.~(\ref{eq:loss-rs});
        \STATE Update task-specific policy $\theta_{\tilde{\mathcal{T}}}$ by SGD:
        $$\theta_{\tilde{\mathcal{T}}} = \theta - \alpha \nabla_{\theta} {\mathcal{L}}_{\tilde{\mathcal{T}}}(\theta);$$
        \STATE Sample $\ell$ extended trajectories $\tilde{\mathcal{D'}}:= \{ \tilde{\tau}'(\tilde{{\mathcal{T}}}) \}$ with meta state embedding $h$, reward shaping $\phi$ and task-specific policy $\theta_{\tilde{\mathcal{T}}}$;
   \ENDFOR
   \STATE Update meta policy $\theta_{\tilde{{\mathcal{E}}}}$ with Eq.~(\ref{eq:loss-rs}) and extended trajectories $\mathcal{D'}$:
   $$\theta_{\tilde{{\mathcal{E}}}} = \theta_{\tilde{{\mathcal{E}}}} - \beta \sum_ {\tilde{{\mathcal{T}}}}\nabla_{\theta} {\mathcal{L}}_{\tilde{\mathcal{T}}}(\theta_{\tilde{\mathcal{T}}});$$
   \ENDFOR
   \STATE Update cross-environment meta state embedding $h$ and environment-specific meta reward shaping $\phi$ with gradient descent by Eq.~(\ref{eq:loss-two});
  \ENDWHILE
\end{algorithmic}
\end{algorithm}

\begin{algorithm}[tb!]
   \caption{New Task Fine-tuning with Hyper-Meta RL}
   \label{alg:fine-tune}
   \begin{algorithmic}[1]
         \STATE {\bfseries Input: } New task ${\mathcal{T}}_{new}$; meta state embedding $h$, reward shaping $\phi$ and policy $\theta$;
         \STATE Sample extended trajectories ${\mathcal{D}}_{new}$ with $h$ and $\phi$;
         \STATE Evaluate $\nabla_{\theta} {\mathcal{L}}_{{\mathcal{T}}_{new}} (\theta)$ using trajectories ${\mathcal{D}}_{new}$;
         \STATE Update task-specific policy $\theta_{{\mathcal{T}}_{new}}$ by gradient descent:
         $$\textstyle \theta_{{\mathcal{T}}_{new}} = \theta - \alpha \nabla_{\theta} {\mathcal{L}}_{{\mathcal{T}}_{new}} (\theta).$$
         \IF{not fixed}
            \STATE Update cross-environment meta state embedding $h$ and environment-specific meta reward shaping $\phi$ with gradient descent by Eq.~(\ref{eq:loss-two})
        \ENDIF
    \end{algorithmic}
\end{algorithm}

\subsection{Theoretical Study and Discussions}\label{subsec:theory}
\noindent{\bf{Policy Consistency.}} For classical reward shaping, any potential-based reward shaping can guarantee the consistency of optimal policy~\cite{ng1999policy,devlin2012dynamic}. However HMRL aims to learn meta modules to fulfill policy adaptation on different tasks rather than getting the optimal policy on a specific task. We are concerned with the question if the meta reward shaping scheme is applied on new task learning, whether the optimal policy keeps still. Our theory shows, using meta reward shaping can maintain policy consistency on new tasks.
\newtheorem{thm}{\bf Theorem}
\begin{thm}\label{thm:policy_consistency}
Assume a new task ${\mathcal{T}} = \{{\mathcal{S}}, {\mathcal{A}}, {\mathcal{P}}, \gamma, {\mathcal{R}}\}$ has the same meta state space ${\mathcal{S}}_m$ with the obtained meta RL model. If the environment-specific meta reward shaping ${\mathcal{F}}(s_{m}, s_{m}')$ is utilized, then it will have the consistency of optimal policy between the extended new task ${\mathcal{T}}' = \{{\mathcal{S}}, {\mathcal{S}}_{m}, {\mathcal{A}}, {\mathcal{P}}, \gamma, {\mathcal{R}} + {\mathcal{F}}\}$ and original task ${\mathcal{T}}$.
\end{thm}
\label{sec: Proof_of_theorem_1}
\noindent{\bf{Proof}}.
Consider the accumulated reward for task ${\mathcal{T}}$, i.e. $\textstyle{G_{\mathcal{T}} = }$ $\textstyle{\sum_{i=0}^{T} \gamma^i {\mathcal{R}}(s_i),}$ where $s_i\in {\mathcal{S}}$. Further the accumulated reward for extended task ${\mathcal{T}}'$ can be calculated as
\begin{eqnarray}
G_{\mathcal{T}'} \!\!\!\!&=&\!\!\!\! {\textstyle{\sum_{i=1}^{T} \gamma^i \left[ {\mathcal{R}}(s_i) + {\mathcal{F}} (s_m^i, s_m^{i+1}) \right]}}\nonumber\\
\!\!\!\!&=&\!\!\!\! {\textstyle{\sum_{i=1}^{T} \gamma^i \left[ {\mathcal{R}}(s_i) + \left( \gamma \phi (s_m^{i+1}) - \phi (s_m^i) \right) \right]}}\nonumber\\
\!\!\!\!&=&\!\!\!\! {\textstyle{\sum_{i=1}^{T} \gamma^i {\mathcal{R}}(s_i) + \sum_{i=1}^{T} \left(\gamma^{i+1} \phi (s_m^{i+1}) - \gamma^i \phi (s_m^i) \right)}}\nonumber\\
\!\!\!\!&=&\!\!\!\! G_{\mathcal{T}} - \phi (s_m^0),\label{eq:equivalent}
\end{eqnarray}which indicates that the difference between the accumulated reward $G_{\mathcal{T}}'$ and $G_{\mathcal{T}}$ is a constant because $\phi (s_m^0)$ is independent with the policy.
Further the goal to calculate the optimal policy for two tasks is to maximize the accumulated rewards $G_{\mathcal{T}}'$ and $G_{\mathcal{T}}$ respectively. 
The optimal policies are equivalent based on (\ref{eq:equivalent}). \hfill$\square$

\section{Experiments}
\subsection{Evaluation Protocol}
\label{subsec: Evaluation Protocol}
In this section, two sparse reward RL applications are chosen from the popular \emph{gym-miniworld} \cite{gym_miniworld} to test the efficiency of the proposed HMRL, i.e., {\em{Hallway}} and {\em{Maze}}.
The reward can be obtained only after each task is finished and in all the two problems, we set the reward as $-{\hbox{used-steps}}$.
Both 1st and 3rd scenarios are studied, which also are considered as different environments\footnote{In experiment section of this paper, scenario is equivalent to environment}.
The agent in the 1st person environment can only obtain the observed information from its own moving direction, while the 3rd environment can obtain the whole scenario observation, e.g., Figure \ref{fig:HallwayProblem} and Figure \ref{fig:MazeProblem}. 
Eventually, 1st person environment is much more difficult than 3rd person environment because of partially observation.

Four baseline meta RL algorithms are compared with the proposed HMRL:
MAML~\cite{finn2017model} (the groundbreaking meta RL algorithm),
NoRML~\cite{yang2019norml} (the state-of-the-art meta RL algorithm for sparse reward setting),
variBAD~\cite{zintgraf2020varibad} (the state-of-the-art Bayesian meta RL algorithm for sparse reward setting) variBAD~\cite{zintgraf2020varibad} and HMRL-w/o-ms (HMRL without meta state embedding, while the potential based reward shaping directly conditioned with different environment states, which is from \cite{zou2019reward}).
To emphasize, MAML is enhanced with the flexible meta-SGD extension technique in \cite{li2017meta}.
All baselines except HMRL-w/o-ms are trained on each scenario independently, because they are not adaptive to the cross environment setting. 
While HMRL-w/o-ms is trained on cross environment setting as the ablation experiments.
For fair comparison, all methods are trained with the same inner-loop and outer-loop iteration numbers.
In the experiment, our policy network contains two layers Convolutional Neural Networks with the kernel size of $[5, 5]$, then we use two fully connected layer with the size of $128$. We use the ReLU activation for all models.
All the baselines use the same policy network. MAML and NoRML's implementations are from \url{https://github.com/google-research/google-research/tree/master/norml}. While variBAD's implementation is from \url{https://github.com/lmzintgraf/varibad}. 
In the variBAD, we use a $128$ unit RNN as the trajectories embedding.

\subsection{Problem I: Hallway}
\label{subsec:Hallway}
\begin{figure}[tb!]
	\centering
	\subfigure[1st Person]{
	\includegraphics[width=0.3\linewidth]{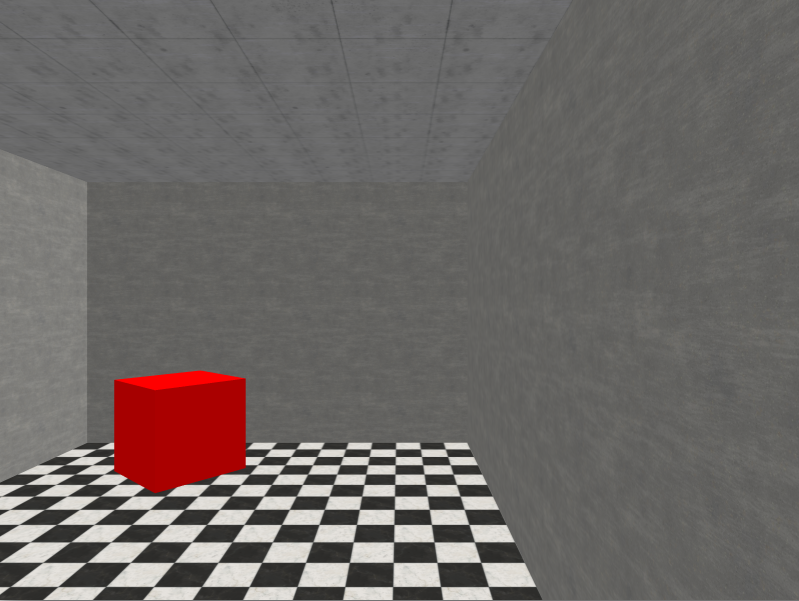}
	\label{fig:first-person-env}
	}
	\subfigure[3rd Person]{
	\includegraphics[width=0.3\linewidth]{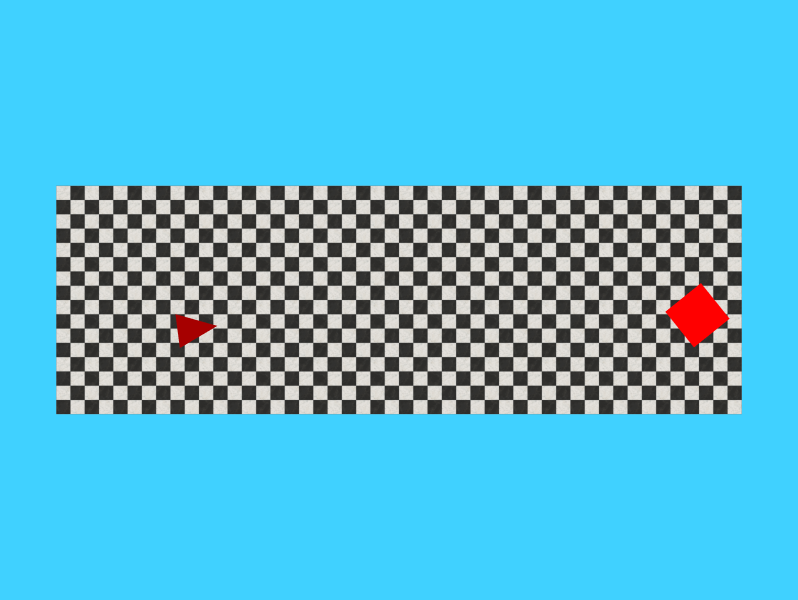}
	\label{fig:third-person-env}
	}
	\subfigure[New task's map]{
	\includegraphics[width=0.3\linewidth]{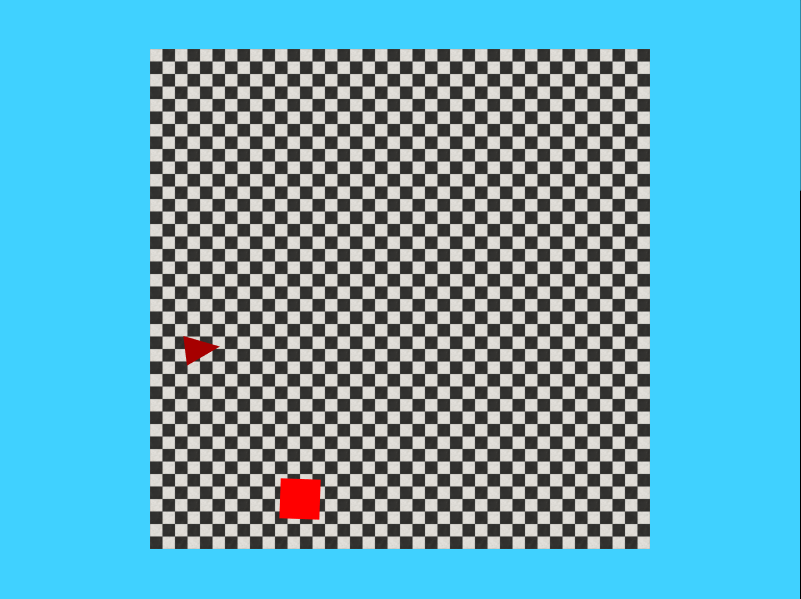}
	\label{fig:test-env}
	}
	\vspace{-10pt}
    \caption{The Hallway Problem. \ref{fig:test-env} shows the scenario/map of the evaluation task, and the task in both 1st person environment and 3rd person environment are evaluated.}
    \label{fig:HallwayProblem}
\end{figure}
\begin{figure*}[tb!]
	\centering
	\subfigure[1st person environment]{
	\includegraphics[width=0.28\linewidth]{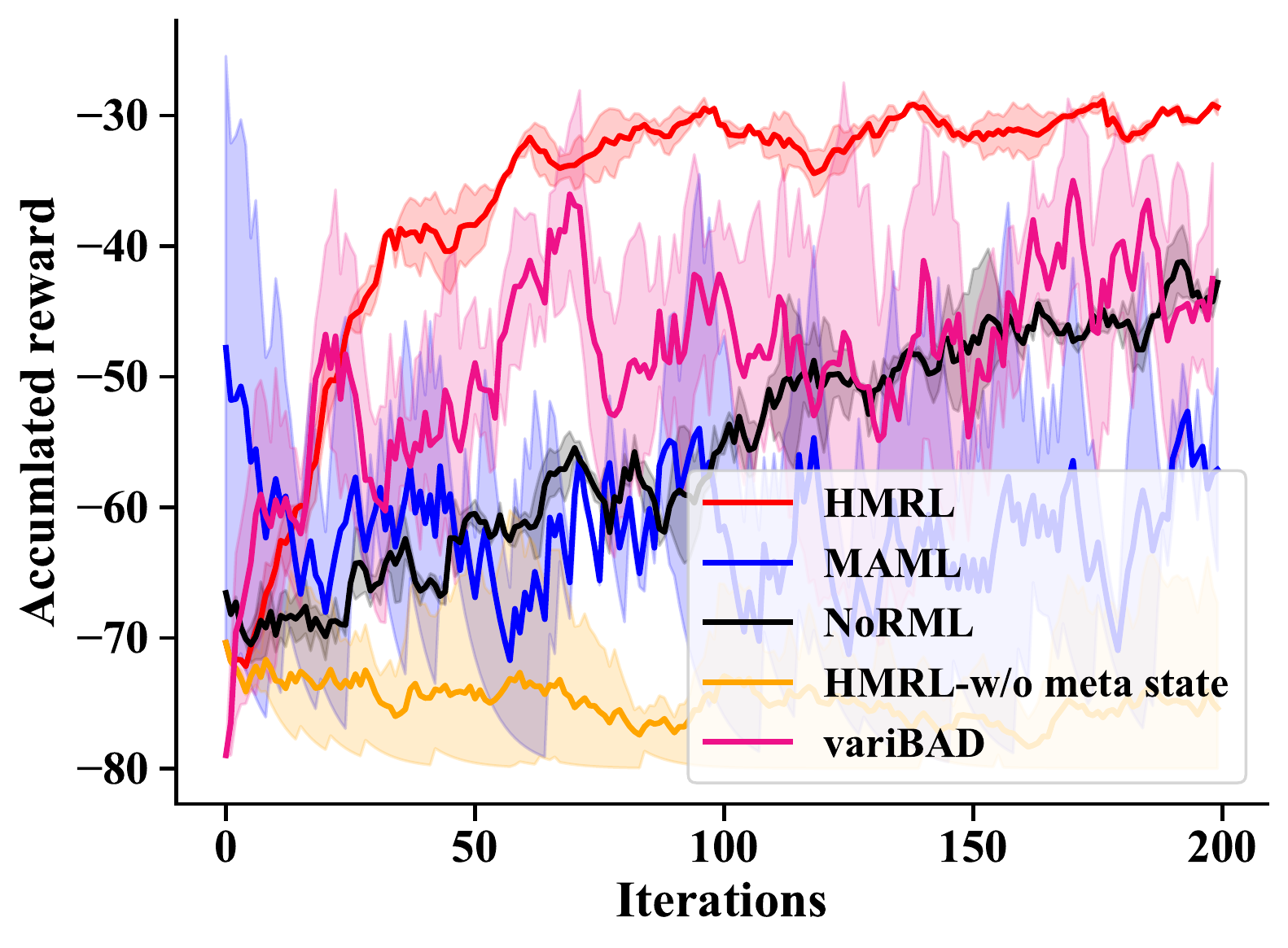}
	\label{fig:learning_curve_3d}
	}
	\subfigure[3rd person environment]{
	\includegraphics[width=0.28\linewidth]{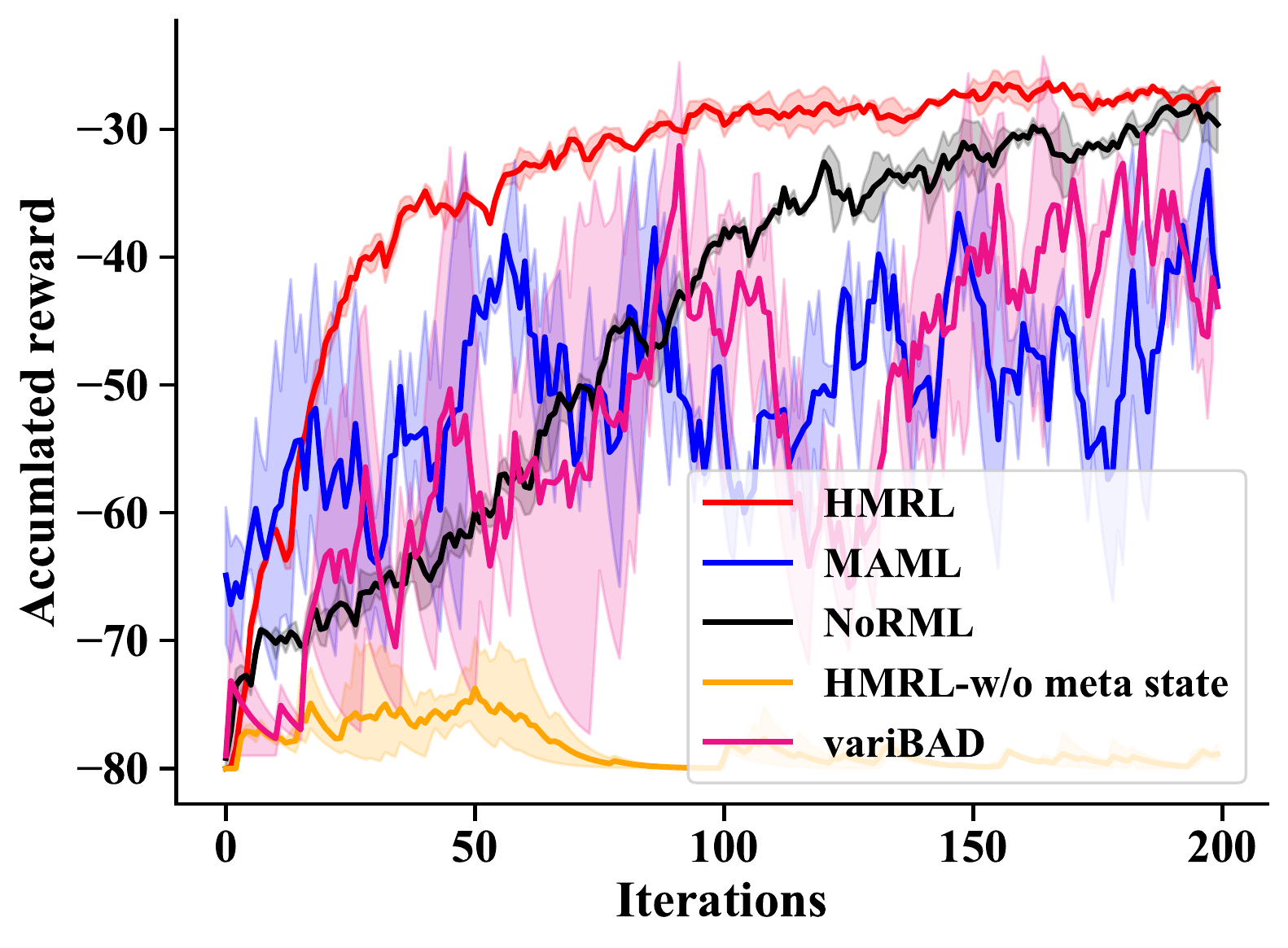}
	\label{fig:learning_curve_2d}
	}
	\subfigure[Used steps on the tasks]{
	\includegraphics[width=0.33\linewidth]{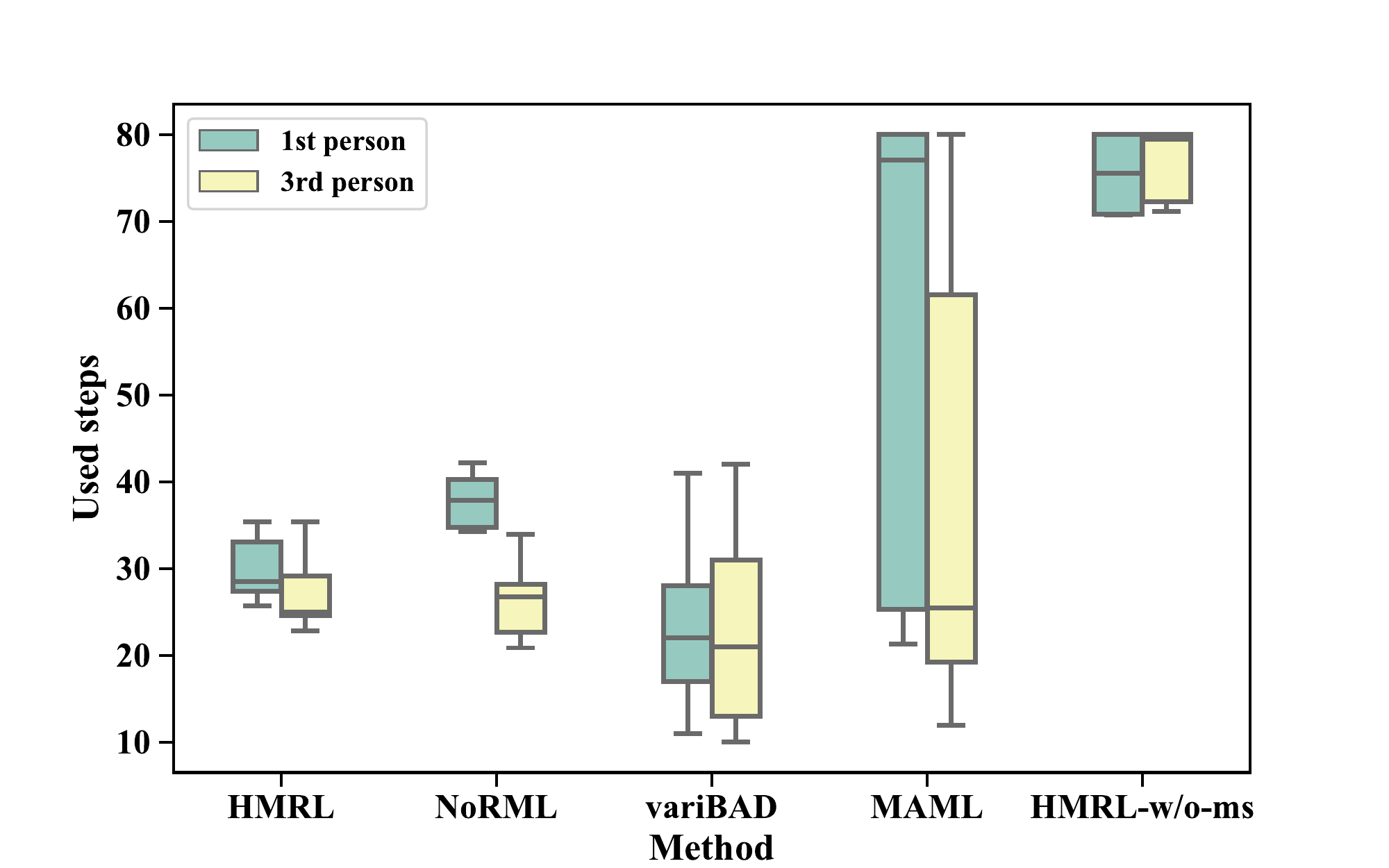}
	\label{fig:test_result}
	}
	\subfigure[Potential value heatmap]{
	\includegraphics[width=0.33\linewidth]{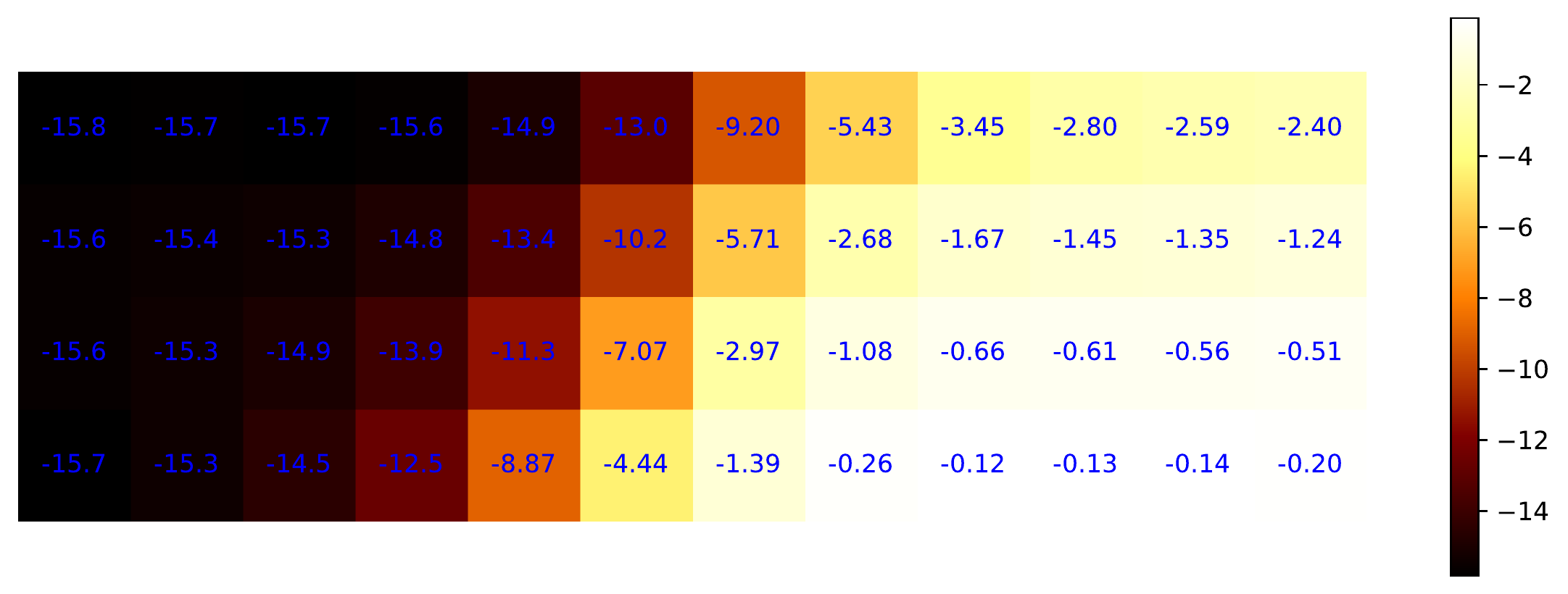}
	\label{fig:Heatmap}
	}
	\vspace{-10pt}
    \caption{Results of the 1st person and 3rd person environments in Hallway problem. \ref{fig:learning_curve_3d} and \ref{fig:learning_curve_2d} shows the learning curve of both environments in training process. \ref{fig:test_result} shows the step numbers for the agent finishes the 10 randomly chosen tasks with the trained policy. \ref{fig:Heatmap} shows the heat map of the averaged potential value in the cross environment meta reward shaping.}
    \label{fig:Learning Curve}
\end{figure*}
The agent aims to reach the red box (or target point) at the end of a straight hallway, while avoiding bumping into walls.
Different tasks in each environment (e.g., 1st person or 3rd person) are defined by the random initialization of the start location.
For both 1st and 3rd person environments, the observation of the agent in each step contains a $80 \times 60 \times 12$ tensor (including recent four frames with RGB channels of its view towards its moving direction, while Figure \ref{fig:first-person-env} and Figure \ref{fig:third-person-env} denotes each frame of two environments respectively) and the coordinates of the agent and the red box.
The action space contains four moving directions
$$\{  {\rm up}, {\rm down}, {\rm left}, {\rm right}\},$$
with $0.2$ unit. The task ends if the agent reaches within $1.0$ distance to the red box and each rollout has a maximum horizon of $80$ steps.

The common meta state $s_m$ is set to be the coordinates of the agent in two environments and the red box, which is denoted as a $3 \times 2$ matrix.
As a result, the meta state embedding function $h$ can be considered as simple concatenation of the location information without learning.
The meta reward shaping potential function $\phi$ is set as a fully-connected network with two $32$ units hidden layers.
The policy network $\pi$ is set as a Convolutional Neural Networks and the setting details are proposed in \ref{subsec: Evaluation Protocol}.

Additionally for evaluation, we introduce a new task with both 1st person and 3rd person environments within a different map, i.e. Figure \ref{fig:test-env}.
The evaluation task is set with
the start point of the agent at $[1.0, 6.0]$ and the red box at $[3.0, 9.0]$.
In order to show the transfer-ability of our algorithm within sparse reward and cross-environment tasks, we further modify a new action space setting:
$$
\{ {\rm turn\ left}, {\rm turn\ right}, {\rm move\ forward}, {\rm move\ back} \},
$$where ``${\rm turn\ left}$" and ``${\rm turn\ right}$" aims to change the agent's direction left and right respectively in a counterclockwise/clockwise direction with $30$ degree, while ``${\rm move\ forward}$" and ``${\rm move\ back}$" aims to move the agent on its current direction with $0.2$ unit.

From Figure~\ref{fig:learning_curve_3d} and Figure~\ref{fig:learning_curve_2d}, our proposed HMRL makes an improvement compared with all other baselines in both 1st person and 3rd person environments.

Compared with HMRL-w/o-ms which directly use the reward shaping without meta state embedding, HMRL can obviously benefit from the cross-environment knowledge. While directly using reward shaping will introduce more noise from heterogeneous state of different environments.

Figure \ref{fig:test_result} presents the step numbers for the agent arriving at the target red box from $10$ randomly chosen tasks, and our proposed HMRL uses the minimum steps in the worst task on 1st environment and achieves a decent holistic performance.
HMRL can promote agent in relatively harder case (e.g., 1st person environment compared with 3rd person environment) exploring more efficiently by the utilization of cross-environment meta information.
Figure \ref{fig:Heatmap} shows the heat map of the averaged potential value in the cross environment meta reward shaping on a specific training task.
The targets (red boxes) are located on the right of the map, and the agent will obtain a positive reward shaping if it moves from the smaller potential value to the larger one.
The heat map proves that the learned cross-environment meta reward shaping encourages the agent moving to the target (red box).

If we proceed further, we can find the learning curves of most baselines shows that training in 1st person environment(Fig.\ref{fig:learning_curve_3d}) is more difficult than 3rd person environment(Fig.\ref{fig:learning_curve_2d}) at the early time because of the low sample efficiency caused by sparse reward and partial observation. But HMRL can bring the effective prior from 3rd person environment to 1st person environment through the cross-environment part. So it represented the training process with a smaller gap between the two environments.

All the evidence above confirmed that HMRL can reach our main motivation of using cross environment information to solve sparse reward problem in meta reinforcement learning because it assuredly learned the cross environment information and the learned cross environment information positively help the meta reinforcement learning tasks with sparse reward problem.
\begin{figure*}[htb!]
	\centering
	\subfigure[1st person environment]{
	\includegraphics[width=0.27\linewidth]{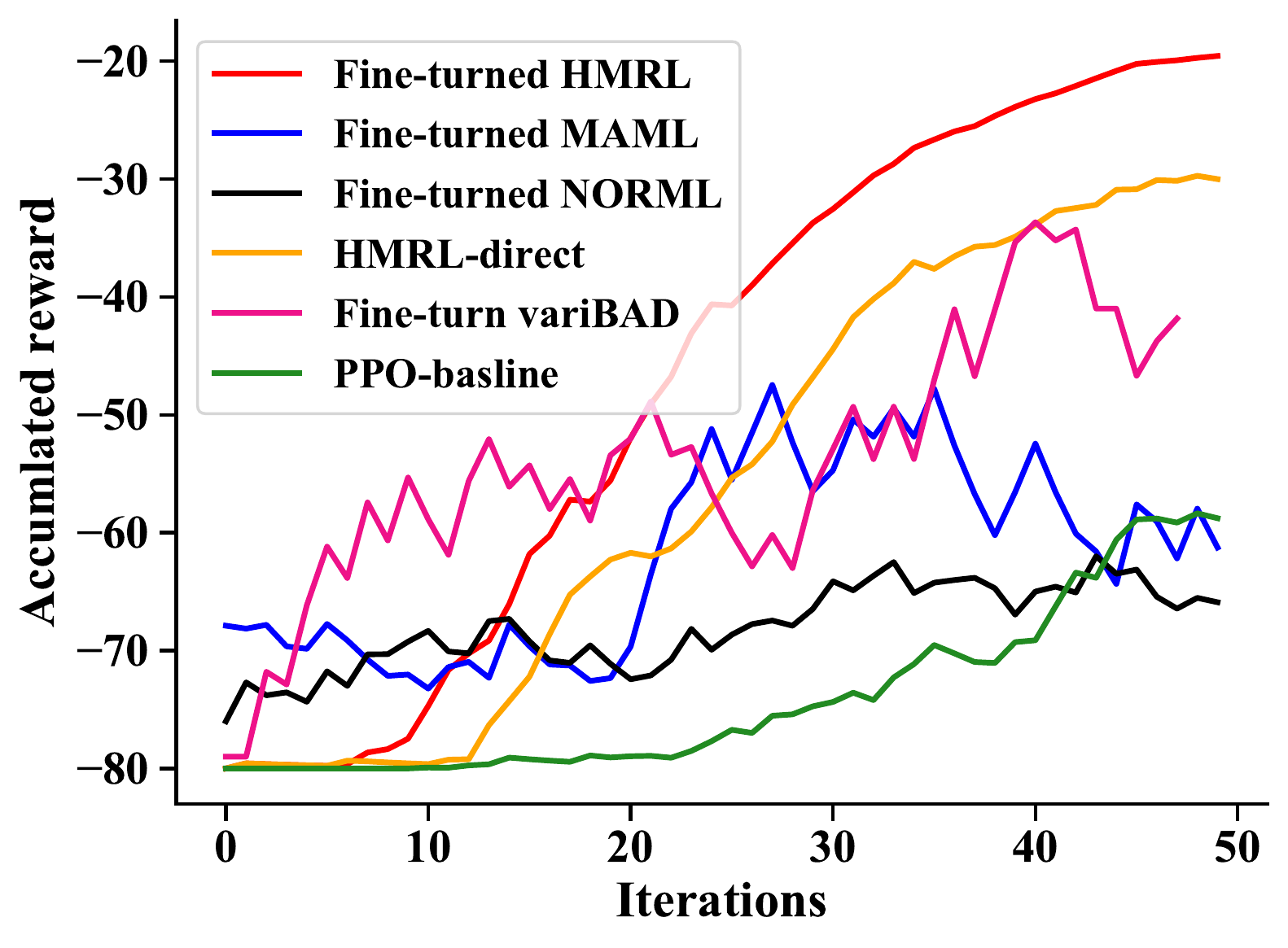}
	\label{fig:test_performance_1}
	}
	\subfigure[3rd person environment]{
	\includegraphics[width=0.27\linewidth]{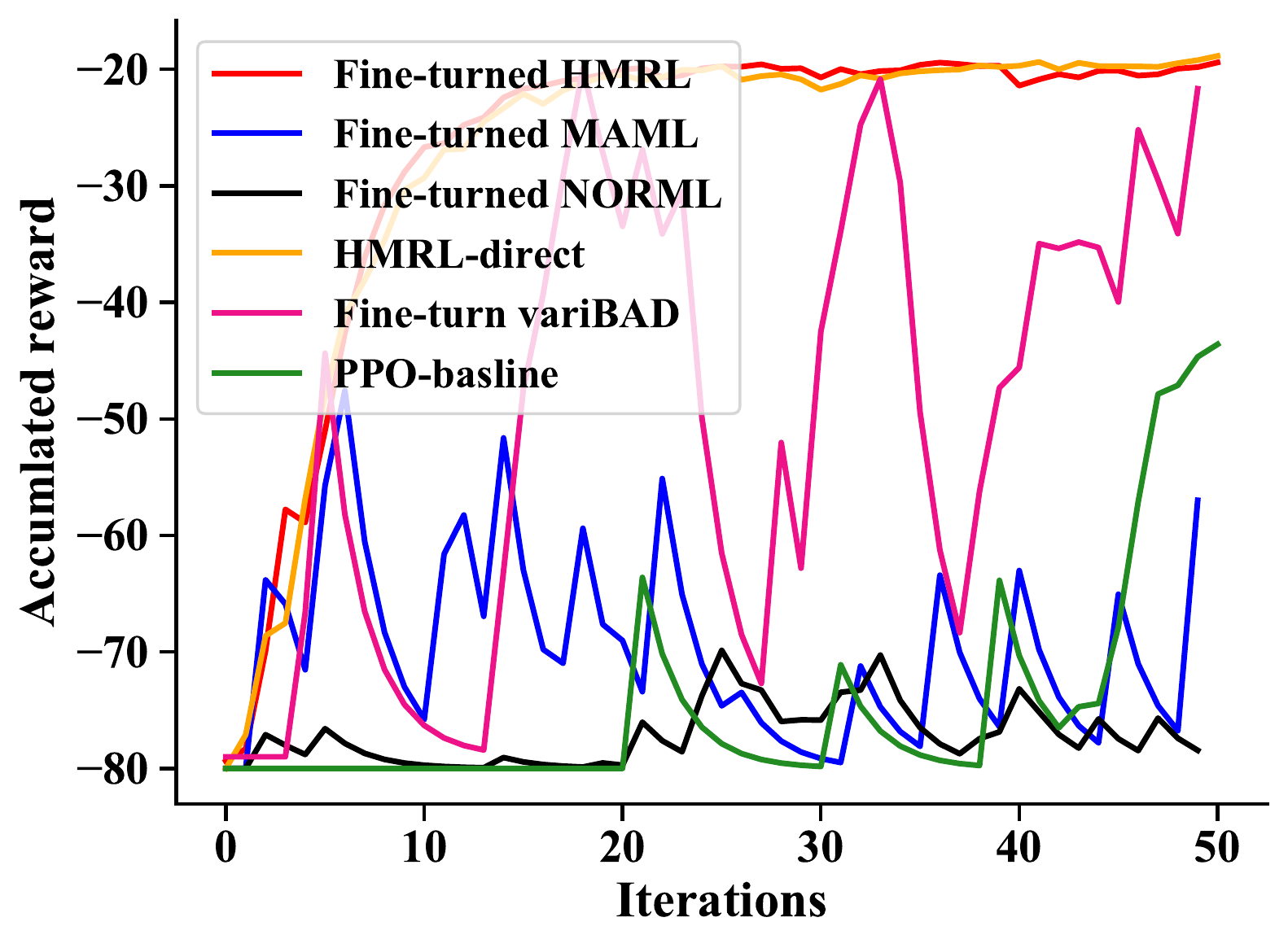}
	\label{fig:test_performance_3}
	}
	\subfigure[${\hbox{HMRL-direct}}$]{
	\includegraphics[width=0.27\linewidth]{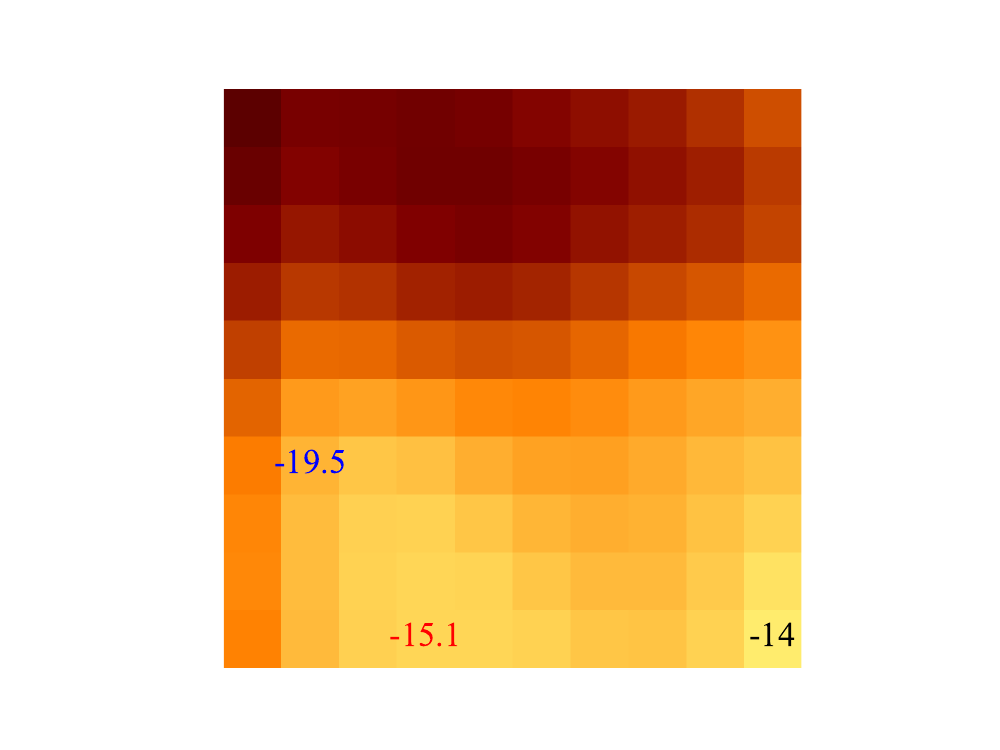}
	\label{fig:DirectHeatMap}
	}
	\subfigure[Fine-tuned HMRL]{
	\includegraphics[width=0.31\linewidth]{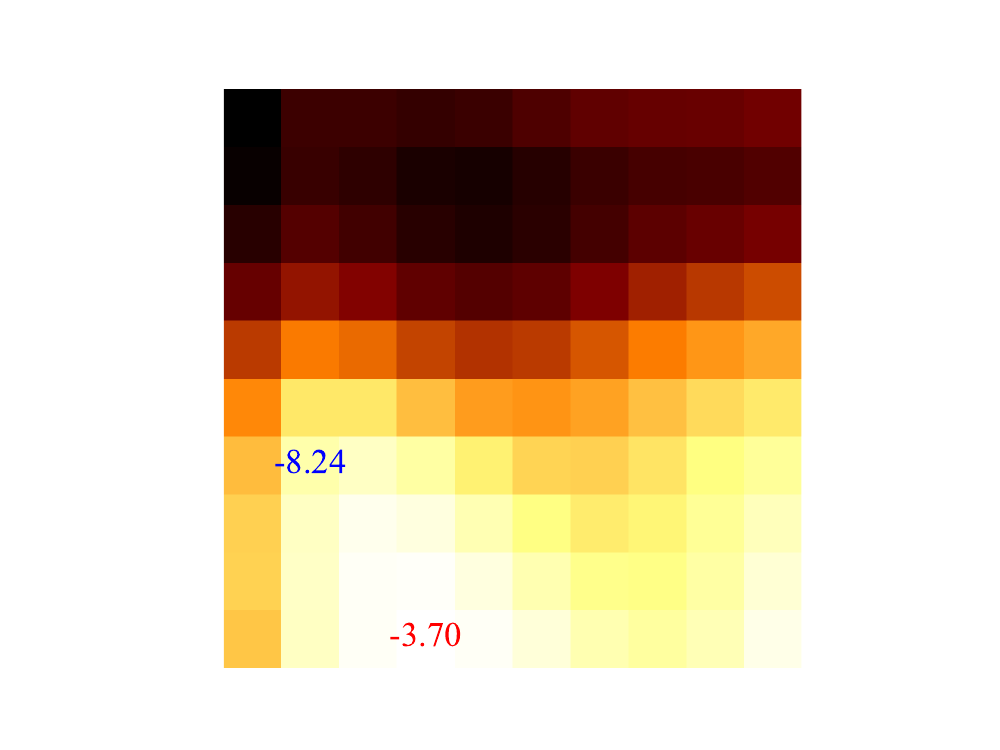}
	\label{fig:FineTurnedHeatMap}
	}
    \subfigure[Trajectories in 1st person environment]{
	\includegraphics[width=0.31\linewidth]{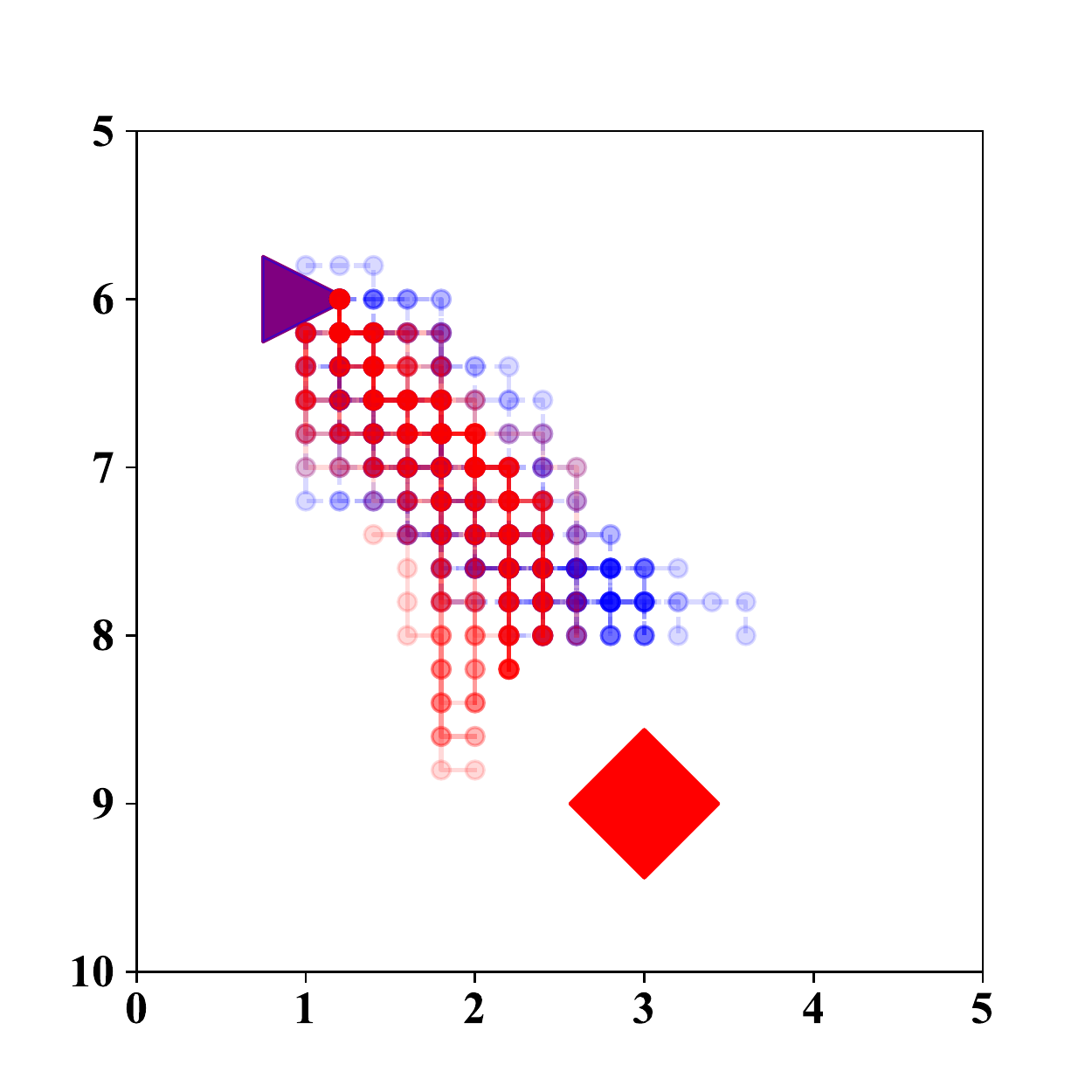}
	\label{fig:traj_1}
	}
	\subfigure[Trajectories in 3rd person environment]{
	\includegraphics[width=0.31\linewidth]{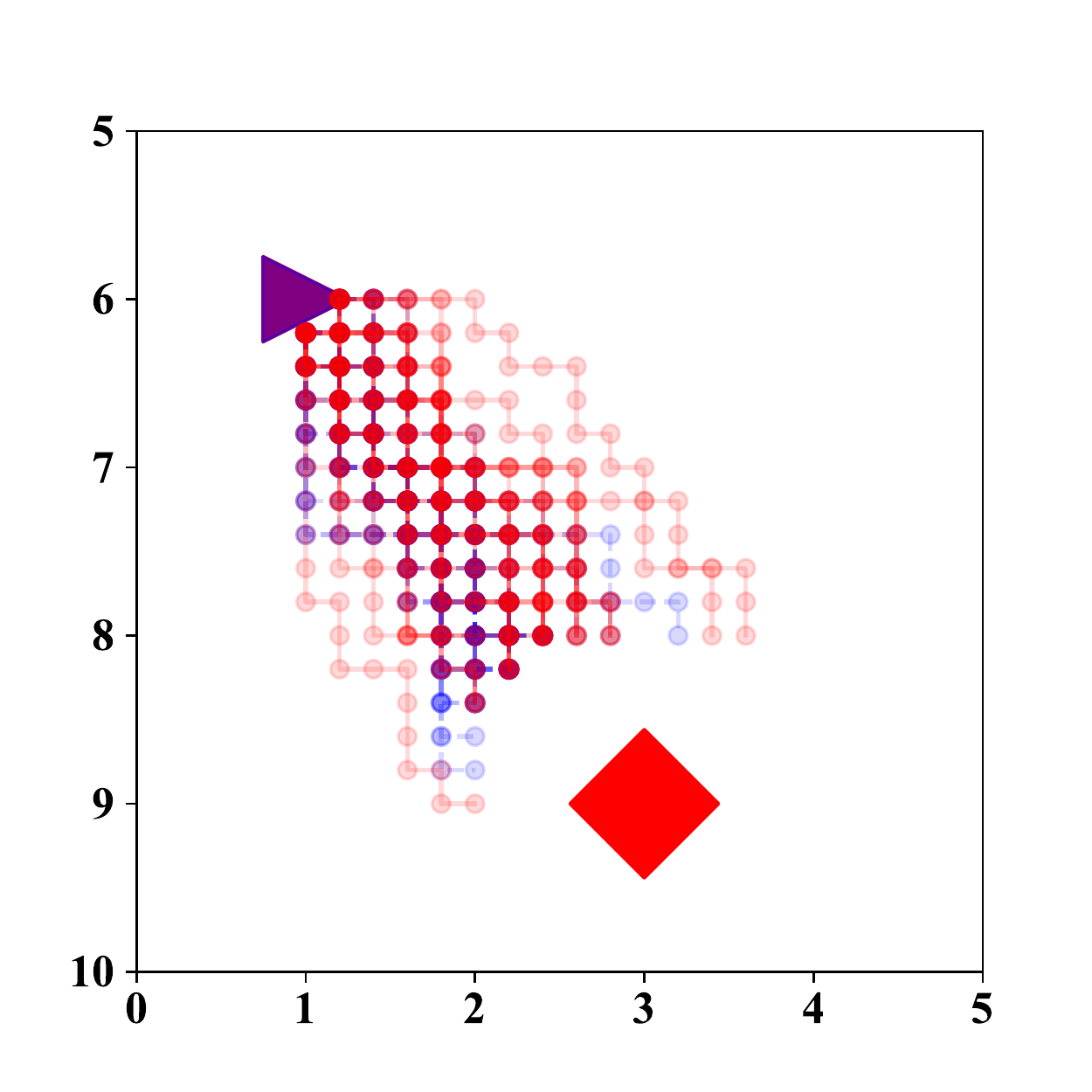}
	\label{fig:traj_3}
	}
	\vspace{-10pt}
    \caption{The results on the new evaluation tasks shown in Figure \ref{fig:test-env} within both 1st person environment and 3rd person environment. The blue, red and black colors denote the ``start" point, ``target" point and ``misleading" point (defined in the main context) respectively. The trangle is the start point and the red box is the target. And the blue trajectories are from the policy trained by HMRL-direct, the red trajectories are from the policy trained by Fine-tuned HMRL. All the trajectories are collected from the policies ran on the corresponding evaluation task for 20 times.}
    \label{fig:AdaptHeatMap}
\end{figure*}

Figure \ref{fig:AdaptHeatMap} shows the result comparison on the evaluation task in Figure \ref{fig:test-env} in order to show the transfer-ability of different algorithms.
The policy obtained by our proposed HMRL can be directly applied to the new task, which is one of the advantages of HMRL and denoted as ${\hbox{HMRL-direct}}$.
All compared algorithms are fine-tuned for better adapting to the evaluation task including the HMRL.
Moreover, the RL method PPO\footnote{All meta learning methods in this paper use PPO as the inner training methods} \cite{schulman2017proximal} is compared and trained from the beginning on this evaluation task as a bottom line.
In Figure \ref{fig:test_performance_1}, we can find that all meta RL algorithms are more efficient than PPO.
Most important, the HMRL outperforms all other baselines on both direct transfer and fine-tuned transfer cases in both environments, while the fine-tuned transfer version performs much better in the 1st person environment.
Figure \ref{fig:DirectHeatMap} and \ref{fig:FineTurnedHeatMap} present the heatmap of the potential value obtained by ${\hbox{HMRL-direct}}$ and fine-tuned HMRL respectively.
In Figure \ref{fig:DirectHeatMap}, the potential value of the target point is not the largest one, as a result we call the point with the ``misleading" point because of the larger potential value usually guiding the agent to move towards.
Conversely, there is no ``misleading" point for the fine-tuned HMRL heatmap, so that the agent have a higher possibility to go on a closer way.
This guess can also be supported by the average trajectories of 1st environment in Fig~\ref{fig:traj_1}, where the policy trained by HMRL-direct guides the agent go to the right-side where the misleading point located. But at the same time, the average trajectories of 3rd person environment in Fig~\ref{fig:traj_3} showes the policy trained by HMRL-direct doesn't have this tendency, and also both HMRL cases have the similar final result in the learning curve in Fig~\ref{fig:learning_curve_2d}. In this point, it seems that Fine-tuned HMRL can get a better performance when treating with new partial observed tasks suffer from sparse reward problem.

Overall, in the experiment, HMRL can jointly utilize the information of tasks from different environments and using the learned cross environment information to speed up and improve both new task learning as well as meta policy training under sparse reward problem. All these figures prove the transferability and policy learning efficiency of HMRL.

\subsection{Problem II: Maze}

\begin{figure*}[tb!]
	\centering
	\subfigure[2RS3]{
	\includegraphics[width=0.22\textwidth]{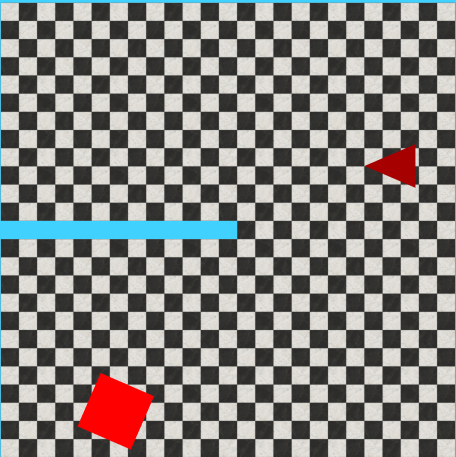}
	\label{fig:2RS3}
	}
	\subfigure[2RS4]{
	\includegraphics[width=0.29\textwidth]{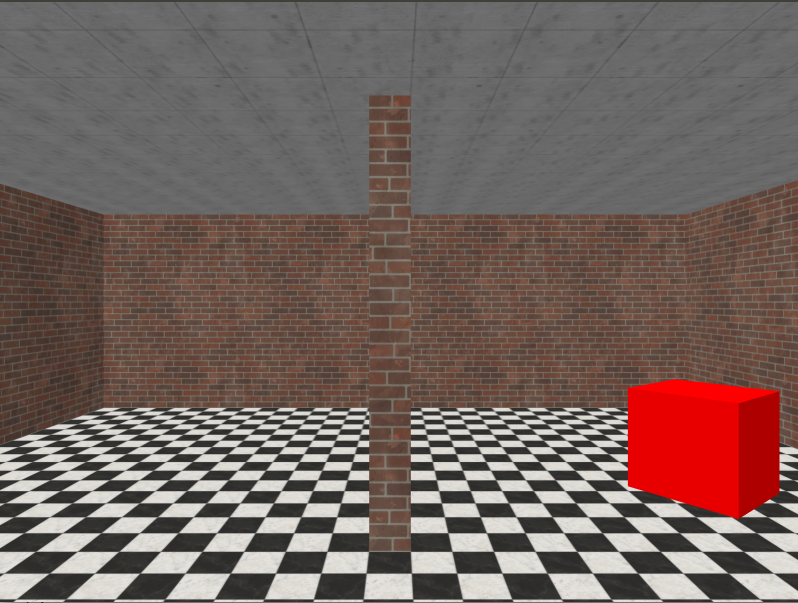}
	\label{fig:2RS4}
    }
	\subfigure[3RS3]{
	\includegraphics[width=0.22\textwidth]{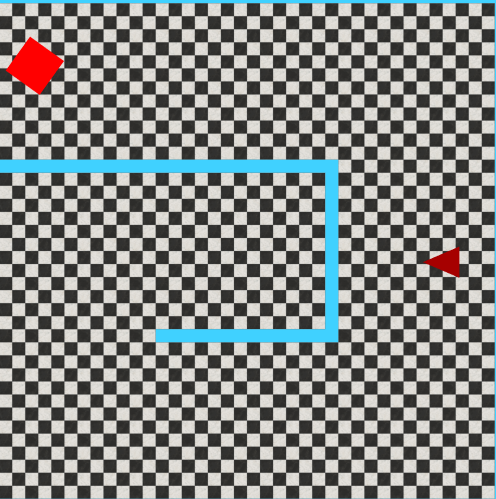}
	\label{fig:3RS3}
	}
		\subfigure[2RS3-heatmap]{
	\includegraphics[width=0.26\textwidth]{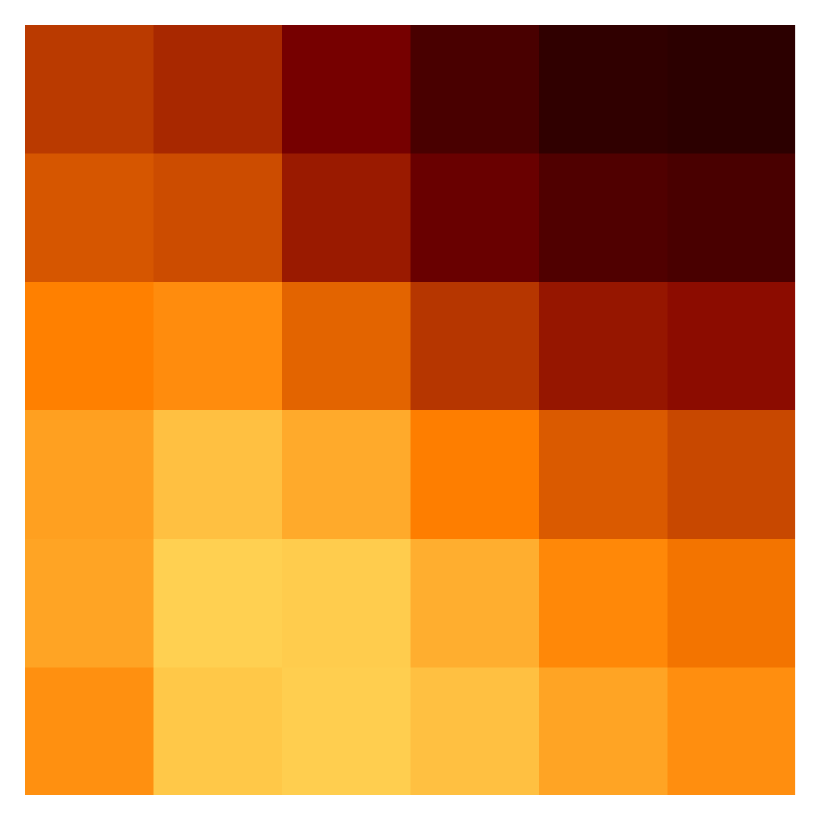}
	\label{fig:2RS3-heatmap}
	}
	\subfigure[2RS4-heatmap]{
	\includegraphics[width=0.26\textwidth]{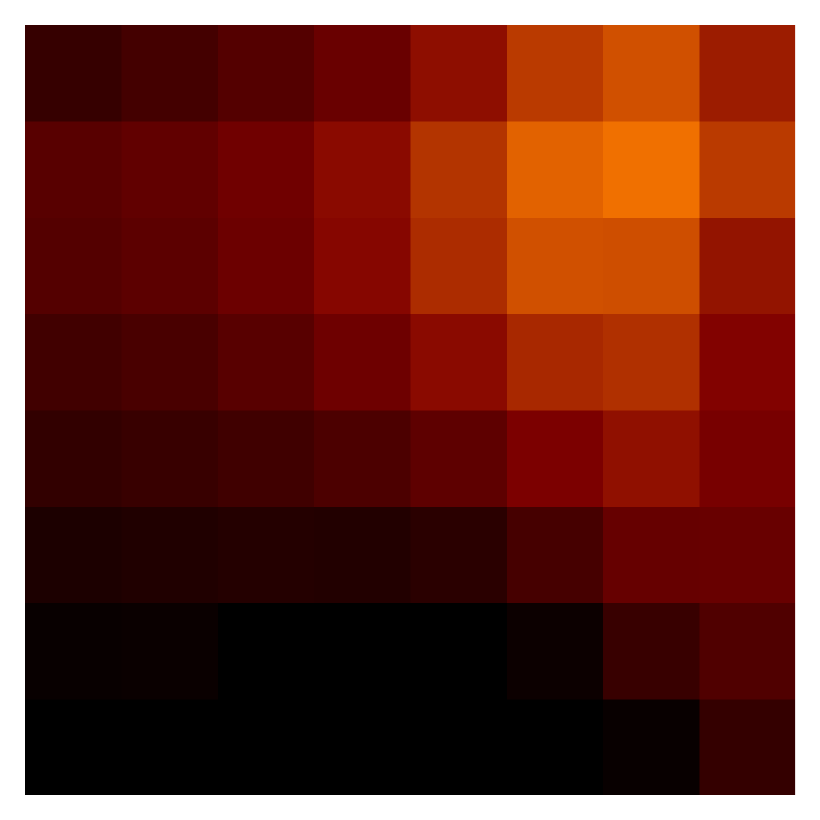}
	\label{fig:2RS4-heatmap}
    }
	\subfigure[3RS3-heatmap]{
	\includegraphics[width=0.325\textwidth]{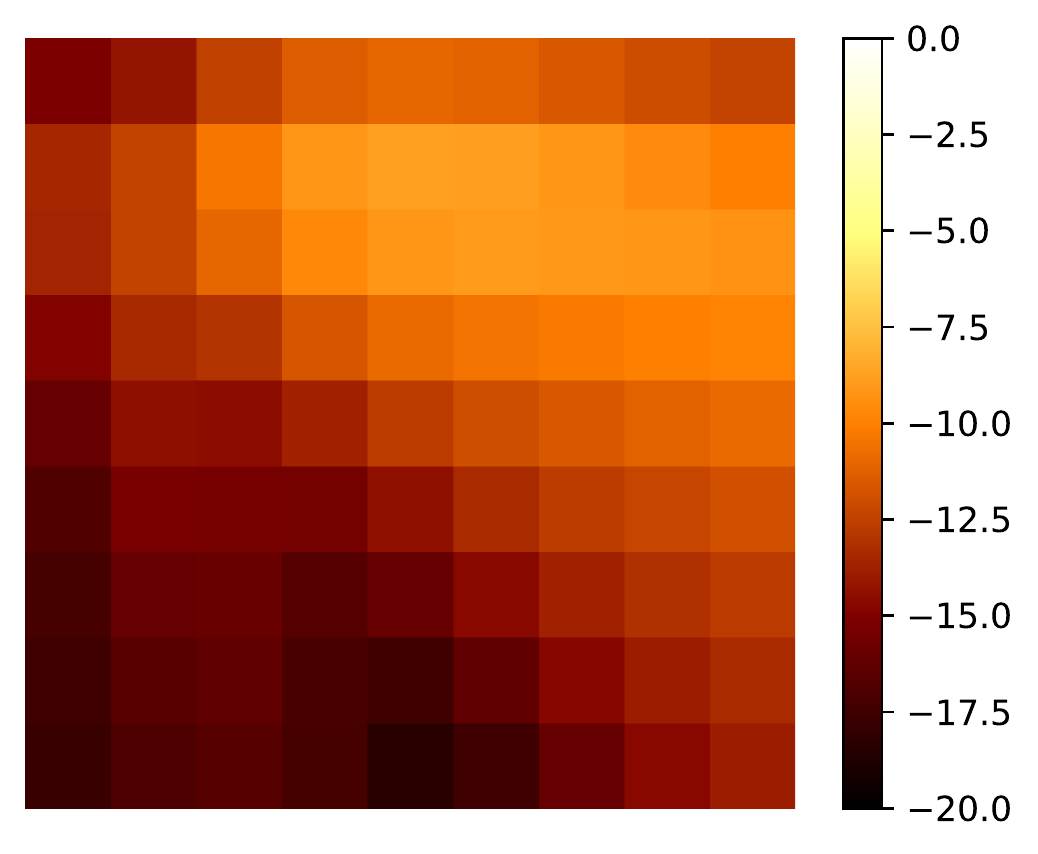}
	\label{fig:3RS3-heatmap}
	}
	\vspace{-10pt}
    \caption{Visualization of Maze Problem and corresponding potential value heatmap. In the chosen task within 2RS4, the red box was located in the upper right.}
    \label{fig:MazeProblem}
    \vspace{-7pt}
\end{figure*}

The agent in Maze problem aims to reach the red box (or target point) in a randomly generated maze.
Three different environments are considered, i.e., \textbf{2RS3}, \textbf{2RS4} and \textbf{3RS3} (Figure \ref{fig:MazeProblem}).
The task for each environment is defined by the random initialization of scenario (both start point and target point).
The brief introduction of these environments are proposed as follows:
1) \textbf{2RS3}: a maze with $2$ rooms and the size of each room is $3$ (shown in \ref{fig:2RS3}) which is called {\em{2R(room)S(size)3}} with \textbf{2RS3} for short.
The observation is obtained in the 3rd person view, and the action space is set to be moving to ${\{{\rm up},{\rm down},{\rm left},{\rm right}\}}$ with a length of $0.2$; 
2) \textbf{2RS4}: a maze with $2$ rooms and the size of each room is $4$ (shown in \ref{fig:2RS4}). 
The observation is obtained in the 1st person view, and the action space is set to be changing the agent's direction left and right ($\{ {\rm turn\ left}, {\rm turn\ right}\}$) and moving to forward or backward ($\{{\rm move\ forward}, {\rm move\ back} \}$) with a length of $0.2$; 
3) \textbf{3RS3}: a maze with $3$ rooms and the size of each room is $3$(shown in \ref{fig:3RS3}), we also set the step length as $0.4$ to speed up the task. The other settings are the same as \textbf{2RS3}.
\begin{figure*}[htb]
	\centering
	\subfigure[2RS3]{
	\includegraphics[width=0.3\linewidth]{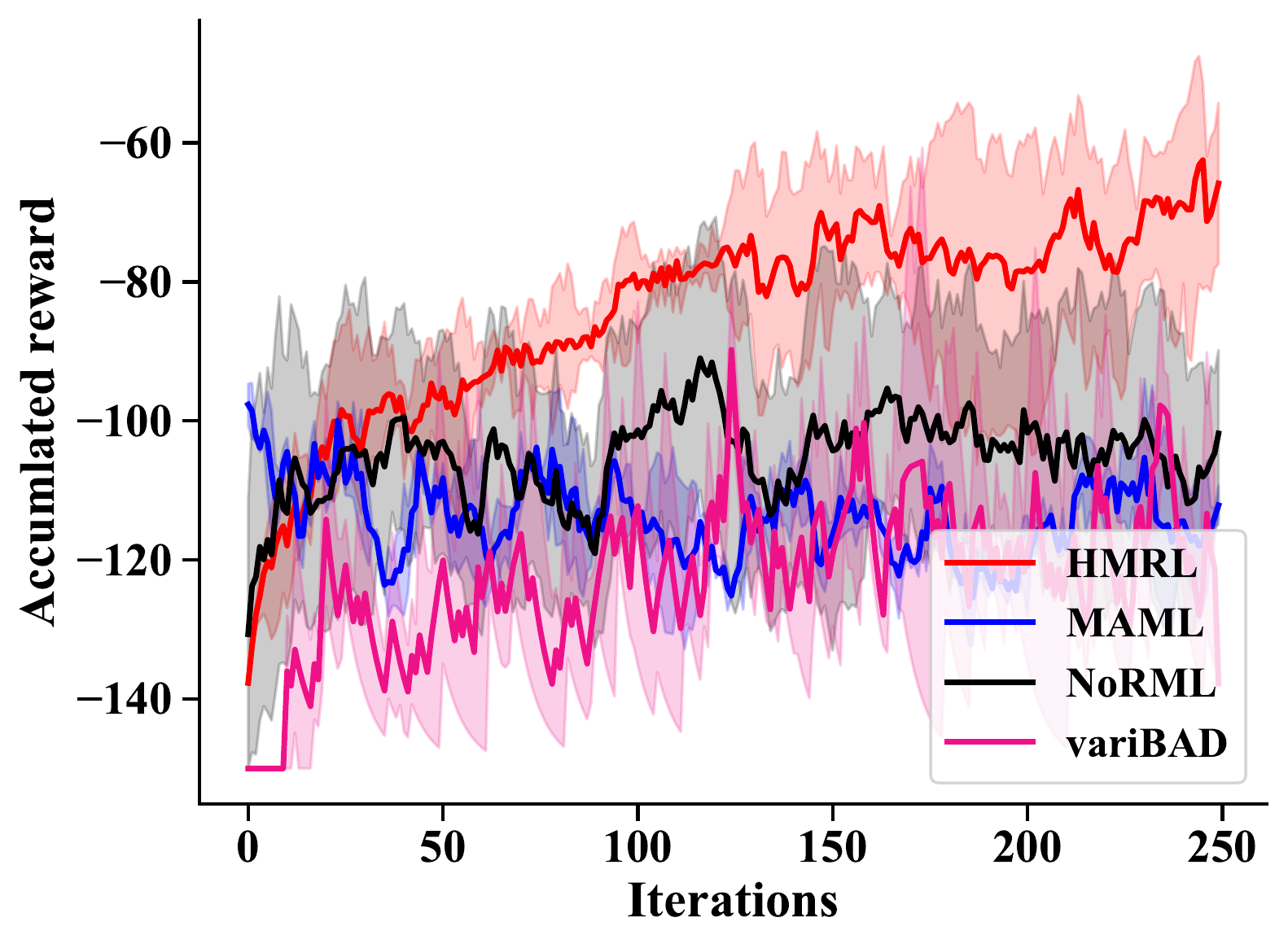}
	\label{fig:2RS3-curve}
	}
	\subfigure[2RS4]{
	\includegraphics[width=0.3\linewidth]{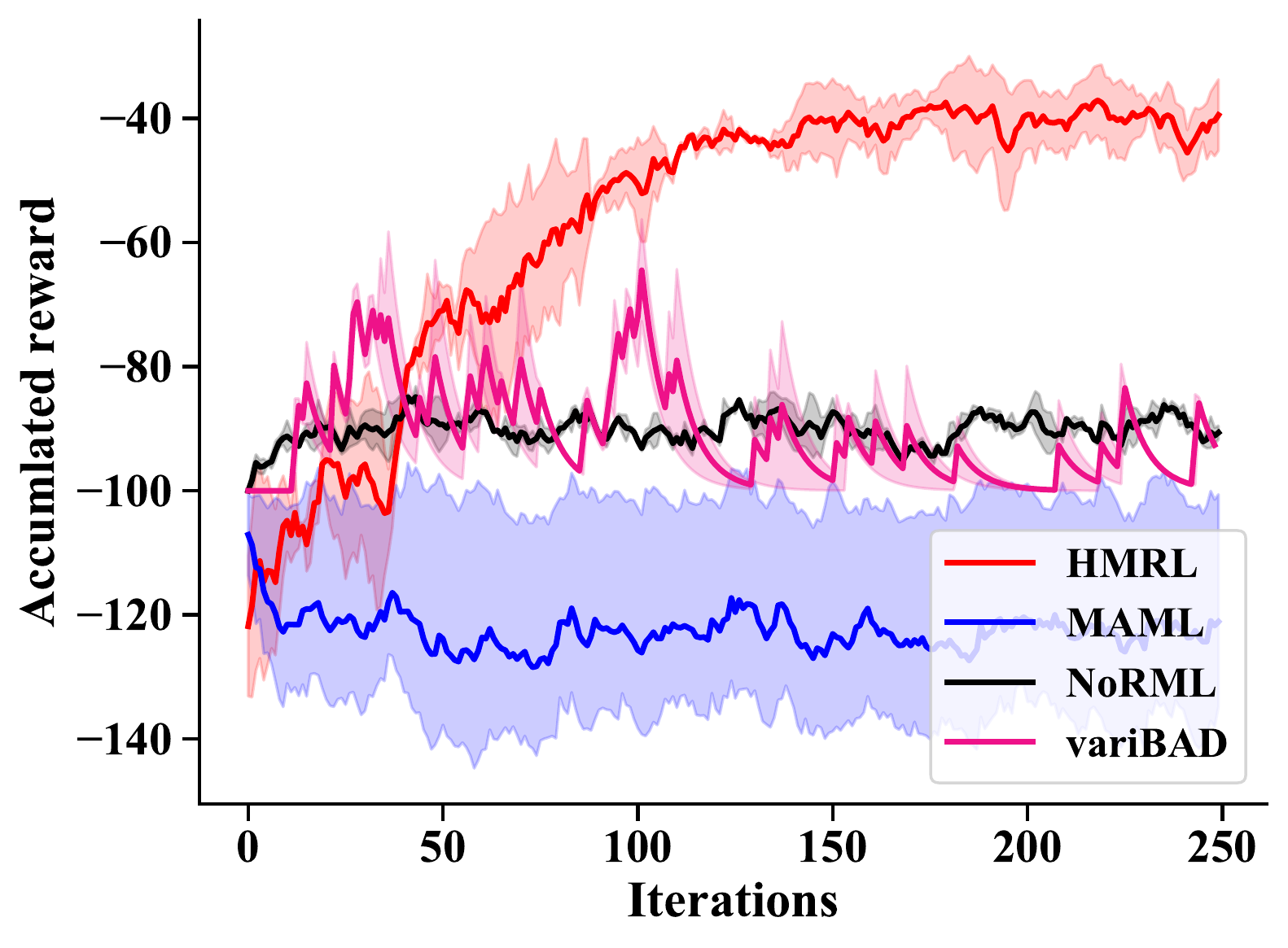}
	\label{fig:2RS4-curve}
	}
	\subfigure[3RS3]{
	\includegraphics[width=0.3\linewidth]{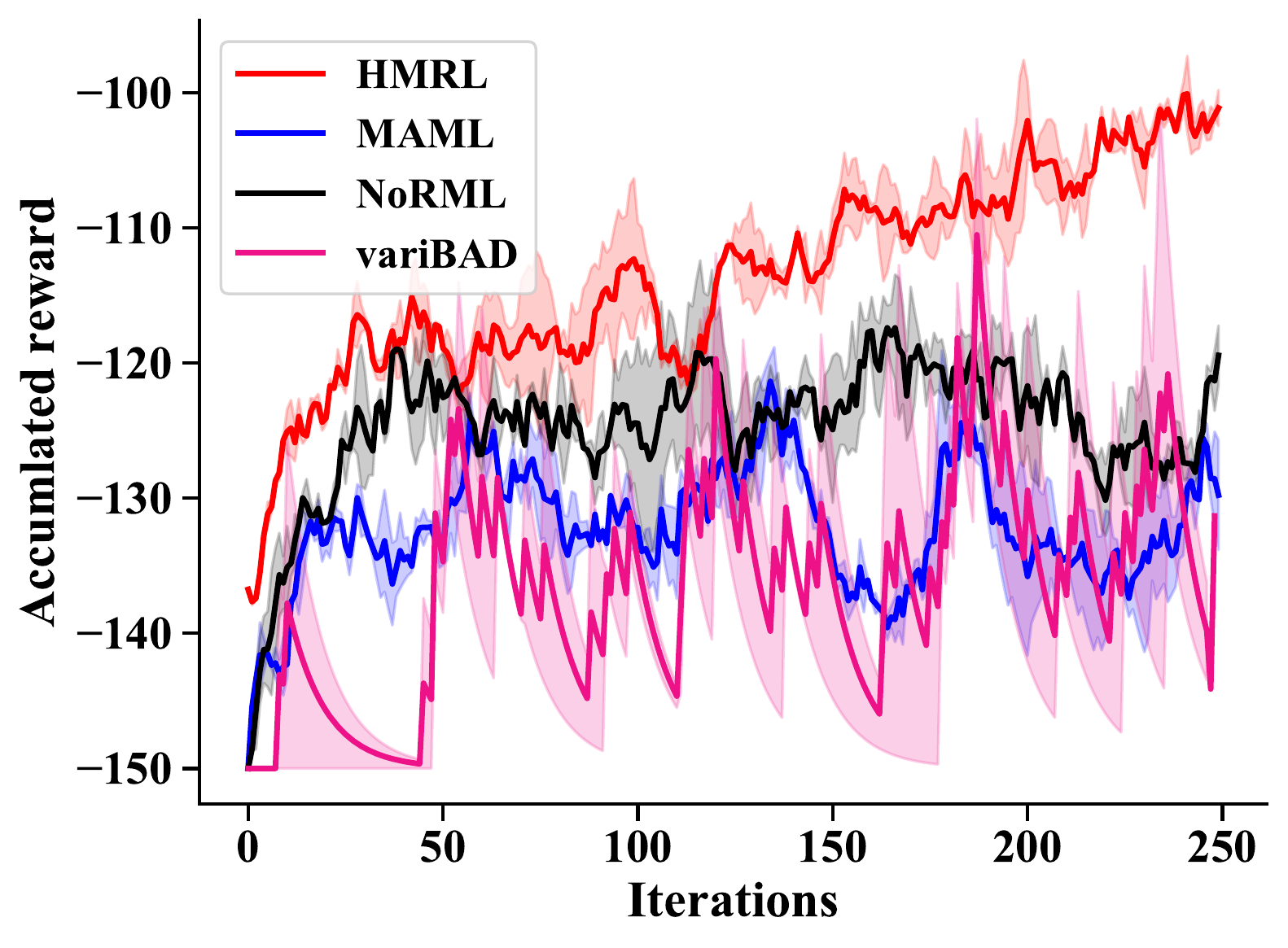}
	\label{fig:3RS3-curve}
	}
	\vspace{-10pt}
    \caption{The learning curve of different environments in Maze problem.}
    \label{fig:Learning Curve of Maze}
\end{figure*}

For all three environments, the observation contains a $80 \times 60 \times 12$ tensor (including recent four frames with RGB channels and the coordinates of the agent and target red box).
Additionally, the task will end if the agent reaches within $1.0$ to the target red box and each rollout has a maximum horizon of $150$ steps.
Due to the difference among these environments, directly setting the common meta state $s_m$ to be the coordinates of the agent and the red box is useless.
For this reason, we set the state embedding function $h$ to be a linear function based on each environment's scenario size (denoted as the domain knowledge $z$), with a $3 \times 2$ mini-map as the common meta state $s_m$.
The meta reward shaping potential function $\phi$ is set as a fully-connected neural network with two $32$ units hidden layers.
The policy network $\pi$ is set as a Convolutional Neural Networks and the setting details are proposed in \ref{subsec: Evaluation Protocol}.
The learning curves are shown in Fig.~\ref{fig:Learning Curve of Maze}.

It is obvious that our HMRL performs better than other baselines in all environments.
And in Figure~\ref{fig:2RS3-heatmap}, Figure~\ref{fig:2RS4-heatmap} and Figure~\ref{fig:3RS3-heatmap}, the potential values on different environments' meta states are proposed.
The meta reward shaping can lead the agent to the true target red box, for the reason that the closer to the target leads to the larger potential value.
So the evidence above indicates that HMRL can transfer meta knowledge among multiple environments with more efficient reward shaping. Furthermore, it shows that in this experiment with a different setting in cross environment meta reward shaping and scenarios than the experiment in \ref{subsec:Hallway}, our HMRL can also assuredly learn the cross environment information from more different environments and using the cross environment information to help training the meta reinforcement learning tasks with sparse reward problem.

In this experiment, HMRL still can utilize the cross environment information in a more general setting.
So that we can say this experiment further prove the transferability and policy learning efficiency of HMRL.


\section{Conclusion}
A hyper-meta reinforcement learning algorithm (HMRL) has been proposed to handle the typical sparse reward problem. 
The cross-environment meta state embedding and environment-specific meta reward shaping modules are devised to cooperate with the general meta policy learning module, which increase the generalization and learning efficiency of meta-RL. Furthermore, a shared meta state space is designed with a specific meta reward shaping technique, which enables and improves the capability of extracting complementary meta knowledge, introducing domain knowledge, speeding up the policy learning for cross-environment tasks. Experimental results with sparse reward demonstrate the superiority of the proposed method on both transfer-ability and efficiency. Our work can also be considered as an early search of designing intrinsic reward in cross-environment problems.

In the future, we will focus more on searching optimal reward in more flexible cross-environment problems. As recent attempts on cross-environment situations are limited in action space and state space, we have taken a small but pioneering step forward. Also, we will explore more challenging environment changes (perhaps by collaborating with other party to access the real-world platforms) whereby our method may have much potential. The source code will be made public available.

\small
\bibliographystyle{plain}
\bibliography{main.bbl}

\newpage

\end{document}